# Can Conversational AI loosen Us-Versus-Them Boundaries? The Effects of Common, Dual, and Separate Identity Framings on Pro-Immigrant Intergroup Helping

Oluwadamilola Jeboda[1], John F. Dovidio[2], & Jonas R. Kunst[3*]

1. Department of Psychology, University of Oslo
2. Department of Psychology, Yale University
3. Department of Communication and Culture, BI Norwegian Business School

*Please direct correspondence to Jonas R. Kunst, Jonas.r.kunst@bi.no

## Abstract

Rising immigration has intensified intergroup tensions in many countries. Traditional bias-reduction programs remain difficult to scale and increasingly constrained by U.S. policy. This preregistered experiment tested whether conversational AI can shift how majority-group members categorize and relate to Latine immigrants. Drawing on the common ingroup identity model, a quota-representative national sample of 658 non-Latine White U.S. adults completed five rounds of dialogue with a large language model (GPT-4o). The model was instructed to frame Latine immigrants in terms of a common ingroup identity (a shared American identity), a dual identity (both Latine and American), or a separate identity (distinct cultural boundaries), or to discuss an unrelated topic in a control condition. The manipulations altered categorization: relative to control, common ingroup identity and dual identity conversations lowered separate categorization, and dual identity conversations raised dual categorization. Although direct effects on behavior and pro-diversity beliefs were nonsignificant, willingness to act was significantly higher in the conditions emphasizing a superordinate identity (common ingroup and dual identity). A path model further revealed indirect associations: both conditions reduced separate categorization, which in turn correlated with greater willingness to act. Semantic similarity analyses of the transcripts confirmed that conversations tracked their assigned narratives; participants' convergence with shared-identity language related positively, and with separate-identity language negatively, to willingness to act. These effects were largely consistent across moderators (need for closure, openness to experience, and political orientation). The findings show that brief AI conversations can loosen us-versus-them boundaries while underscoring the gap between cognitive recategorization and behavior.

*Keywords:* artificial intelligence, common ingroup identity, dual identity, separate identity, immigration attitudes, intergroup bias, large language models, prosocial behavior, recategorization

**Can Conversational AI loosen Us-Versus-Them Boundaries? The Effects of Common, Dual, and Separate Identity Framings on Pro-Immigrant attitudes and Intentions**

Immigration rates continue to increase globally. While this may be favorable for increasing positive intergroup contact, labor supply through employment, and labor demand through entrepreneurship (Azoulay et al., 2022; Krol, 2021), the corresponding shift in demographics can elicit more negative attitudes toward certain immigrant groups (Huo et al., 2018). While attitudes toward undocumented immigrants are especially negative, immigrants who reside in the country legally also experience prejudice and discrimination. Negative untrue claims, often expressed in political campaigns (Hickel & Bredbenner, 2020), arouse perceived threats to the ethnic majority culture, raise questions about who is deserving of citizenship, and exacerbate bias against immigrants (Esses, 2021; Udani & Kimball, 2018). These negative attitudes and stereotypes can fuel discriminatory policies and personal actions that restrict opportunities for immigrants, which not only unfairly disadvantage immigrants, but also members of society who would benefit from the resources that immigrants bring to the country (Esses & Sutter, 2026).

Much of the immigration discourse in the United States has been focused on Latine immigrants specifically, representing the largest foreign-born population in the United States. Analyzing group-specific mentions of Latine people in prominent newspapers from the 1980s to the early 2000s, Valentino et al. (2013) found that coverage was relatively low between 1985 and 1993. But by 2000, direct mentions of Latine people in immigration-related stories were significantly higher than other groups, and by the mid-2000s, Latine people were the most mentioned group by far. Additionally, Valentino et al. (2013) found that when White people think of immigrants, Latine people come to mind most readily, and participants as a whole (from a national sample) were most worried about Latine immigrants' effect on communities in the United States (as compared to East Asian, African, and Eastern European immigrants), illustrating the increased salience of Latine immigrants in U.S immigration discourse and opposition to it.

To address these negative perceptions and barriers to social cohesion, psychological research has traditionally focused on bias-reducing interventions, such as anti-bias training (Devine & Ash, 2026), that is typically administered in one or more sessions to small groups of participants. The present research investigated how recent technological developments in artificial intelligence (AI) can deliver scalable, theory-based interventions to improve attitudes toward and support for immigrants. It had two main objectives. The first was to apply AI chatbot technology as a research tool to influence the way non-Latine American participants socially categorize Latine immigrants, which creates a psychological foundation for intergroup orientations generally (Kawakami et al., 2017). The second objective was to investigate the direct and indirect pathways from the chatbot induced social categorization to prosocial intentions and behavior to support Latine immigrants.

Anti-bias training (also sometimes called diversity training) represents a family of programs "aimed at facilitating positive intergroup interactions, reducing prejudice and discrimination, and enhancing the skills, knowledge, and motivation of participants to interact with diverse others" (Bezrukova et al., 2016, p. 1228). However, evidence of the effectiveness of these interventions, in part because of their diffuse objectives or limited grounding in psychological theory, has been mixed, and traditional anti-bias training has been criticized for not justifying the significant financial investment that supported these programs (Dobbin & Kalev, 2016). Anti-bias programs have also been substantially dismantled in the United States through a series of Presidential Executive Orders and federal policies beginning in 2025, based on arguments that these programs were harming traditional American values. The present research adopted an approach that differs from anti-bias training. Its framework, the Common Ingroup Identity Model (Gaertner & Dovidio, 2000), drawing on extensive psychological theory and research, can be tailored to address unfair treatment of members of a particular group, and does not violate current policies that restrict diversity, equity, and inclusion training in the United States.

## The Common Ingroup Identity Approach to Reducing Intergroup Bias

To mitigate hostility toward immigrants and foster more equitable treatment, we leverage psychological mechanisms identified in the Common Ingroup Identity Model (Gaertner & Dovidio, 2000; see also Dovidio & Kunst, 2026). This model, which is an established framework for achieving prejudice reduction, builds on principles integral to Social Identity Theory (Tajfel & Turner, 1979). Social Identity Theory posits that individuals automatically socially categorize others based on salient characteristics, distinguish between those who belong to the same group as the individual (the ingroup) or to another group (an outgroup), tend to identify with their ingroup, and desire for their group to be positively distinguished from others (Billig & Tajfel, 1973). Ingroup bias, in which people spontaneously value and act more favorably towards members of their group over members of other groups, is hypothesized to arise from the desire of individuals for their social group to be positively distinguished from others (Turner et al., 1979). Subsequently, people are willing to act more favorably towards ingroup members (Kawakami et al., 2017). When negatively invoked, these social categorization and social identity processes can lead to an “us versus them” mentality and treating outgroup members less favorably. However, social categorizations of who is an ingroup member and who is an outgroup member are malleable and continuously refined through social comparison as new interactions and information emerge.

Inducing individuals to broaden definitions of whom they consider to be an ingroup member promotes more favorable and equitable orientations to those persons now viewed as belonging to the same group (Gaertner et al., 2015). The Common Ingroup Identity Model describes the process of reducing intergroup bias in which separate ingroup and outgroup categorizations are reconfigured around a common, inclusive identity (Gaertner & Dovidio, 2000). Interventions that change the way groups are socially categorized from separate to one-group identity, extend the forces of ingroup favoritism to those previously considered to be outgroup members. Viewing others more strongly only in terms of their separate, outgroup identity leads to greater bias, whereas increasing one-group identity fosters more positive intergroup attitudes and promotes intergroup helping behaviors (Gaertner & Dovidio, 2000; Gaertner et al., 2015).

**One-Group and Dual Categorization**

A common identity refers to a sense of solidarity and belonging within a unified social group (e.g., “We are all Americans”). The effects of a common ingroup identity on intergroup bias have been studied extensively and shown to influence both attitudes and behaviors between members of the majority group and members of a minoritized group (Charnysh et al., 2015; Kunst et al., 2015; see also Dovidio & Kunst, 2026, for a review). The process of recategorizing people from being an outgroup member to one-group categorization can encounter challenges. Because of historical or current nature of relations between the groups (see Perez et al., 2026), people may be resistant to incorporating members of particular groups within a shared identity. For example, the Ingroup Projection Model posits that ingroup members regard prototypical aspects of their primary group identity as positive and normative, while outgroup attributes are seen as a violation of superordinate category norms (Wenzel et al., 2007). Thus, recategorization into a superordinate identity can lead to ethnocentrism as majority-group members define what a typical group member (i.e., "American") is like and derogate deviations from this ideal. In the same vein, work on the "irony of harmony" (Saguy et al., 2009) suggests that one-group categorization can promote positive intergroup attitudes but also may undermine motivation to act prosocially on behalf of those who are discriminated against (Banfield & Dovidio, 2013). Lastly, members of another group may be reluctant to replace a valued, distinctive group identity with a common ingroup identity. In such cases, adopting dual identity may be more optimal for promoting positive attitudes and behaviors between groups.

In a dual categorization (e.g., "U.S. American and Latin American"), a common superordinate identity, and a subgroup identity are simultaneously salient. This categorization can also achieve the significant benefits of shared identity compared to conditions in which separate group identities are salient, or to control conditions, improving both attitudes and behaviors towards former outgroup members (Kunst & Thomsen, 2015; Verkuyten, 2007). Additionally, dual identity orientations preserve distinct subgroup identities and are thus preferred by members of minoritized groups (Hehman et al., 2012).

**The Potential of Using AI Chatbots to Scale Recategorization**

The impact of the superordinate identity approach for improving intergroup relations has been constrained by resources and limited scalability. The advent of generative AI, and in particular chatbot interfaces to communicate with large language models (LLMs), has changed that. Some research has shown that these LLMs are highly persuasive in changing deep seated social beliefs (Costello et al., 2024).

The bias reduction interventions utilized in previous studies have largely relied on static, one-way communication, such as asking participants to read fixed texts (e.g., newspaper articles emphasizing different identities, common identity, or dual identities with respect to the participant's ingroup and another group; e.g., Banfield & Dovidio, 2013). These passive manipulations often have only modest psychological impact. By contrast, active engagement through communication with others (a "saying is believing" effect) generally produces stronger and more enduring impact on people's beliefs (Echterhoff et al., 2005). To address the limitations of traditional, passive stimuli and build on the saying is believing effect, we turn to emerging technologies that offer dynamic, personalized interaction. Modern advances in artificial intelligence have made LLMs a unique tool for engaging with, and challenging individuals' beliefs with depth and personalization. The ability of these language models to draw on vast, diverse information to directly counter or support specific reasoning interactively makes them an ideal tool for eliciting wide ranging viewpoints and potentially changing attitudes.

LLM chatbots have an advantage in that they can engage participants in multi-turn dialogue, addressing concerns and providing counterarguments in real time (Costello et al., 2024). This adaptivity is well suited for challenging the wide range of potential attitudes and beliefs that may be held across a large number of participants. Chatbots may be perceived as more credible and less biased than traditional sources such as news articles, as explained by the "machine heuristic," a mental shortcut in which people believe that machine- or algorithmically-generated information is more likely to be objective and unbiased (Sundar, 2008). This perception grants chatbots a degree of credibility and trustworthiness when presenting new or counter-attitudinal arguments. For example, Zarouali et al. (2021) found that participants agreed significantly more with a counter-attitudinal news story about immigration when it was provided by a chatbot compared to an online news website, also rating the chatbot as more credible. People are also more willing to engage with chatbots on sensitive topics, trusting machine agents more than human agents for personal information disclosure (Sundar & Kim, 2019) and preferring human-computer interfaces when discussing negative views or sensitive topics (Pickard et al., 2016). One reason why human-computer interfaces are preferred is that individuals perceive less judgment and have more time to formulate responses compared to interactions with humans (Pickard et al., 2016). Thus, chatbots combine persuasive credibility with reduced social evaluative threat, making them particularly suitable tools for challenging existing beliefs and encouraging recategorization, even among those who hold strong views.

Previous research employing human-computer interfaces suggests the particular promise of this technique for studying a range of social phenomena. Costello et al. (2024) used an AI chatbot to persuade participants against their belief in a particular conspiracy theory, and found that following just three rounds of conversation, participants' conspiracy belief was reduced by 19.4% as compared to 2.9% in the control condition. This effect extended to those who indicated that their conspiracy belief was at the highest level of importance to their worldview, and the effect remained undiminished two months after the AI chatbot intervention (Costello et al., 2024). LLMs are also effective in political persuasion. Hackenburg et al. (2025) found that LLMs were at least as effective as humans (if not more) in right/left leaning persuasive messaging related to political issues such as vaccine mandates and immigrant deportations. In the context of messaging that was counter to participants' political orientation (Democrat/Republican), the LLM was more effective in persuading right-leaning participants against their beliefs about these political issues, while there was no significant advantage for left leaning counter persuasion (Hackenburg et al., 2025).

Building on this capacity for persuasion, the current study deploys an LLM to operationalize the theoretical frameworks of the Common Ingroup Identity Model through conversation. In the context of Common Ingroup Identity, the LLM can be prompted to dissolve boundaries by reframing immigrant integration as a strengthening of the existing national "we," emphasizing shared goals and values. Conversely, in the Dual Identity condition, the model can validate the distinct cultural heritage of Latine immigrants while anchoring them within the superordinate American identity, framing difference as a complementary asset rather than a hindrance. To test the potential for harm, a Separate Identity condition can utilize the same adaptive capabilities to reinforce rigid boundaries, emphasizing cultural incompatibility and distinctness. Crucially, this approach differs from traditional text-based interventions in its use of responsive recategorization. Unlike static texts that present a fixed set of arguments, the chatbot uses tailored counterarguments to address the specific concerns and hesitations of the participant.

## The Present Research

Majority group members often have the advantage of greater resources and reduced resistance when enacting social change; yet challenges to mobilizing this group to support minoritized groups persist (Kunst & Dovidio, 2026). Thus, the primary aim of the present study was to determine whether a scalable, AI-based recategorization intervention can induce dual and one-group categorizations among non-Latine White majority-group members in the United States and thereby increase their willingness to help immigrants. We focus specifically on the attitudes and intentions of non-Latine White participants regarding Latine immigrants. As reviewed earlier, this minoritized group has become increasingly salient in contemporary media and political rhetoric, often serving as a target for polarizing discourse. We utilize national identity as the superordinate category for this investigation because it can be inclusive by definition (based on citizenship or residence rather than race) yet can paradoxically form the foundation for xenophobia when narrowly defined (Hatungimana, 2024; Taniguchi, 2021).

**Tests of the Overall Effects of the AI Chatbot Recategorization Intervention**

In the present experiment, non-Latine White Americans engaged with an AI chatbot instructed to effectively persuade participants to categorize Latine immigrants as one group, dual groups, or separate groups. These experimental conditions are compared against a Control condition, in which the conversation centered on an unrelated topic – technology in daily life. This design not only allows us to test bias reduction strategies but also whether such technology can be weaponized to deepen social divisions via the Separate Identity condition.

We tested a series of pre-registered predictions. Based on evidence that chatbot conversations can effectively change attitudes, we predicted that AI conversations would alter the social categorizations of non-Latine White Americans and immigrants (H1). Specifically, we expected that conversations emphasizing the benefits of a Common Ingroup Identity would lead to increased one-group categorization and decreased separate categorization relative to control, as the arguments focus on common interests between U.S. citizens and Latine immigrants. Similarly, conversations emphasizing dual identity were predicted to lead to increased dual categorization and decreased separate categorization relative to control, as the chatbot highlights the compatibility of subgroup distinctiveness within a superordinate whole. At the same time, we tested for categorization spillover effects; for example, the possibility that the Dual Identity condition may also lead to more one-group categorization or the Common Ingroup Identity condition to more dual categorization due to the conceptual relatedness between holding a dual identity and acknowledging a shared superordinate group. Conversely, we predicted that separate identity conversations would lead to increased separate categorization and decreased dual and one-group categorizations relative to control, driven by arguments reinforcing distinct "us versus them" boundaries.

We also examined the downstream effects of the intervention on beliefs and behaviors. Based on findings that inclusive identities foster more favorable views of outgroup members (Charnysh et al., 2015; Kunst et al., 2015; Verkuyten, 2007), we predicted that Common Ingroup Identity and Dual Identity conversations would lead to higher pro-diversity beliefs relative to control, whereas Separate Identity conversations would decrease pro-diversity beliefs (H2). Drawing on the Common Ingroup Identity Model's assertion that inclusive categorization motivates prosocial behavior, we further predicted that Common Ingroup Identity and Dual Identity conversations would increase intergroup helping intentions and behavior relative to control (H3). Separate Identity conversations were expected to have the opposite effect, reducing helping intentions and behaviors by reinforcing psychological boundaries that maintain intergroup bias and limit prosocial motivations across group lines.

**Potential Mechanisms of AI-Based Recategorization**

The current research also investigated potential mediators and moderators of the impact of the AI chatbot identity manipulations. The effects on intergroup helping behaviors may operate through cognitive categorizations or changes in attitudes. When outgroup members are recategorized as ingroup members through common ingroup identity or dual identity framings, they become targets of ingroup favoritism and benefit from the same prosocial motivations that individuals direct toward fellow ingroup members (Gaertner & Dovidio, 2000). Under separate categorization, outgroup members remain subjects of intergroup bias and reduced helping motivation. Social identity theory posits that self-categorization drives intergroup behavior, as individuals are motivated to favor those who they deem to be fellow ingroup members over outgroup members to positively distinguish their group from others (Turner et al., 1979). Shifting these categorizations of minoritized outgroup members (in this case Latine immigrants) can influence intergroup helping behavior, overcoming previous intergroup biases without directly targeting attitudes or beliefs about outgroup members (Charnysh et al., 2015).

By contrast, the attitudinal pathway suggests that recategorization can be achieved by changing beliefs and evaluations about diversity and immigrant groups. When individuals adopt inclusive identities, they develop more positive views of diverse groups in society, consistent with previous research linking positive attitudes towards immigrants to support for integration policies and helping intentions (Kunst & Thomsen, 2015). In terms of this pathway, instilling a superordinate identity may succeed by persuading individuals that diversity is valuable, which motivates them to act prosocially toward immigrant groups.

These two pathways are distinct, but they may operate simultaneously. The cognitive categorization pathway emphasizes the direct effect of how people represent group memberships on intergroup behaviors. Individuals are motivated to help because they recognize mutual group membership, viewing former outgroup members as part of "we/us." The attitudinal pathway emphasizes evaluation-based motivation. Individuals are motivated to help because they value diversity and its impact on society. The cognitive categorization pathway may produce helping behaviors without attitude change, purely through categorization processes, while the attitudinal pathway requires a conscious re-evaluation of outgroup members. The present study examined both mechanisms in relation to Latine immigrants and determine the extent to which they contribute to observed intervention effects.

To understand the mechanisms driving these behavioral changes, we tested a mediation model comparing a cognitive route (via categorization) to an attitudinal route (via a shift in pro-diversity beliefs; Kauff et al., 2019) (H4). Pro-diversity beliefs are assessed in the present research with a scale that measures individuals' beliefs that diversity is beneficial for the progress of society.

Common Ingroup Identity condition effects on helping were expected to be mediated by increased one-group categorization (and potentially dual categorization in the case of spillovers) and reduced separate-group categorization, as shifting the cognitive boundary to include the outgroup facilitates helping. Similarly, the Dual Identity condition was predicted to indirectly increase helping via increased dual categorization and reduced separate categorization. With respect to the attitudinal route, we expected that both Common Ingroup Identity and Dual Identity conditions would increase helping via the mechanism of pro-diversity beliefs involving immigrants, because viewing the outgroup as part of the ingroup (or a valued subgroup) improves majority-group attitudes towards minoritized-group members. Conversely, the Separate Group condition was expected to decrease intergroup helping by increasing separate categorization and discouraging inclusive categorizations (cognitive route) or lowering pro-diversity beliefs (attitudinal route) relative to the control condition.

**Potential Factors Moderating AI-Based Recategorization**

We further examined how individual differences in ideology and personality would moderate AI intervention effectiveness. The potential for conversational recategorization (with a chatbot or otherwise) may be influenced by factors including political orientation, need for cognitive closure (Webster & Kruglanski, 1994), and Openness to Experience (Lee & Ashton, 2018). Each of these measures has been shown to relate to social orientations generally; we further investigate whether and how they may moderate recategorization processes.

Understanding these moderating factors is critical for determining who is most likely to respond to recategorization and for tailoring interventions accordingly.

With respect to political ideology, liberals tend to be more in favor of immigration, while those holding a more conservative political ideology tend to have stronger anti-immigrant views and to display less willingness to support immigrants (Kiehne & Ayón, 2016; Rambaud et al., 2021). Because people holding conservative political ideologies are generally motivated to justify and maintain dominant cultural norms and hierarchies, preferring tradition and stability over social change (Jost & van der Toorn, 2011) and opposing pluralism (Verkuyten, 2011), we hypothesized (H5) that participants who are more conservative would be more resistant to adopting a shared identity with Latine immigrants in the chatbot condition emphasizing dual identity but be more responsive to endorsing separate identities in the chatbot condition intended to promote the different identity of Latine immigrants. Because more politically conservative individuals more strongly favor assimilation as an acculturation ideology (Verkuyten, 2011), we further posited that they may be more responsive in adopting a one-group appeal in the Common Ingroup Identity condition, which emphasizes national unity and shared values (Hanson et al., 2021).

In addition to political orientation, we examined the potentially moderating influences of two personality-related individual difference factors, Need for Closure (NFC) and Openness to Experience. NFC is conceived as the preference for a firm belief on a given topic as opposed to confusion and uncertainty (Webster & Kruglanski, 1994). In intergroup settings, this can manifest through seeking consensus within ingroups, a tendency towards ingroup bias, and less identification with outgroup members (Shah et al., 1998). In the present research, we drew upon the Roets and Van Hiel (2011) NFC Scale, which distinguishes a variety of different sources of uncertainty (e.g., novel situations, social disagreements). In general, people who have a higher need for closure are more prejudiced toward groups whom they perceive as "outsiders." However, the particular nature of the chatbot social categorization conditions may have an additional influence on participants' responses. The different identity framings vary in cognitive complexity and categorical clarity. We hypothesized (H6) that the condition that would be least effective for producing the intended representation would be the Dual Identity condition. Dual categorization requires maintenance of two group identities simultaneously (e.g. American and Latine), which introduces a degree of ambiguity. By contrast, because high-NFC individuals prefer clear, unambiguous boundaries (e.g. "We are all Americans" or "They are immigrants"), we predicted that participants higher in NFC would be more likely to endorse a one-group categorization in the Common Ingroup Group Identity condition and a separate categorization in the condition emphasizing the groups' differences.

Openness to Experience may also constrain belief change with respect to orientations toward Latine immigrants. In the HEXACO model, Openness to Experience measures a person's willingness to consider new ideas and engage with novel experiences (Lee & Ashton, 2018). Those who score high in openness are more comfortable with complex and novel ideas and, arguably, by extension the ambiguity that are characteristic of dual categorizations. Thus, we hypothesized that participants higher in Openness to Experience would be more likely to accept dual identities as complementary instead of contradictory and would therefore endorse a dual categorization more strongly in the chatbot conversation emphasizing that representation of the groups (H7). Conversely, separate and one-group categorizations may be more appealing to those who score low in Openness to Experience.

Theoretically, the present research advances the Common Ingroup Identity Model by moving beyond text-based manipulations to dynamic AI interactions. By systematically comparing common, dual, and separate identity framings within a responsive and interactive environment, we can better delineate the boundary conditions of recategorization, specifically, whether the cognitive complexity of dual identities acts as a barrier for individuals with specific personality traits, such as high need for closure, or distinct political ideologies. Furthermore, by explicitly modelling the Separate Identity condition as an active intervention, we offer critical insight into the psychological symmetry between the processes of prejudice reduction and prejudice escalation.

Practically, the findings of the present research potentially hold significant implications for social interventions and digital policy. If AI chatbots prove effective at scaling recategorization, they offer a relatively cost-effective, highly personalized tool for promoting social cohesion and intergroup helping that goes beyond the reach and adaptability of traditional interventions. However, the present study also underscores the critical dual-use nature of this technology; demonstrating that AI can effectively deepen divisions through separate identity framing would highlight the urgent need for relevant ethical safeguards in the deployment of LLMs. Ultimately, understanding how political and personality factors moderate these effects will allow policymakers and practitioners to tailor integration messages more effectively, ensuring that efforts to reduce bias resonate with diverse segments of the population.

## Method

The present research was pre-registered at https://osf.io/kxa94/overview?view_only=dd94b3c416534345b4d4a9d6ebc4ffbc. All data and

code to reproduce the results can be found at:

https://osf.io/m2k9c/overview?view_only=c0567ef4e83c4c4c86e246a3ef96b099

**Participants**

A preregistered power analysis using the InteractionPoweR v. 0.2.2package (Baranger et al., 2023) indicated that 800 participants were needed to achieve at least 90% power to detect a two-way interaction of $\beta = .15$ at the .05 significance level in an ANCOVA. The parameters for this power analyses were set by previous research, addressing concepts similar to the present study. Research by Hilbig et al. (2014) observed that Openness to Experience was correlated at 0.17 with helping orientation. Brewer et al. (2023) found that political orientation was correlated at .20-.23 with helping behavior. Costello et al. (2024) reported that the smallest effect of AI conversations on deep-seated political views was $r = .37$ (see pre-registration for how this informed the exact power analysis parameters).

In total, 851 participants were recruited for an “AI Chatbot Conversation Study” through Prolific. Participants were confirmed to be both born in, and residents of the U.S. through a Prolific pre-screening measures and one additional survey item asking whether they were born in the U.S. Pre-screening measures were used to select White participants who were representative of the U.S. population in terms of gender, political orientation, and education. Ethnicity was again confirmed through a survey item, ensuring all included participants were non-Latine White. Because the U.S. Census Bureau only collects data on sex assigned at birth (U.S. Census Bureau, 2024a), the gender quota was supplemented with data from the Pew Research Center (Brown, 2022). Data for political orientation were derived from the Pew Research Center (2025), and the U.S. Census Bureau (2024b) data for individuals in the workforce were used to determine representative sample parameters for education level. Pre-screening included selecting for participants with a 99–100% approval rating and those with 50–100,000 previous submissions.

Moreover, in Prolific, a bot screening app was used to exclude participants relying on agentic AI or otherwise being flagged as inauthentic. No participant failed that test. Participants were paid $9 per hour of their time and compensated through Prolific. After excluding participants based on attention checks described later, 658 participants were included in the final data analysis, falling somewhat below our pre-registered criterion. A power sensitivity analysis indicated that this resulted in a minimum detectable effect of $\beta = 0.16$ at 90% power, which is slightly larger ($\Delta\beta = 0.01$) than the originally pre-registered effect size.

Of the participants who passed the attention checks, scores on political orientation, $M_{\text{political orientation}} = 4.47$, $SD_{\text{political orientation}} = 3.35$, indicated that participants were slightly left leaning. In terms of education, 0.9% of participants had some high school or less ($n = 6$), 14.1% had a high school diploma or GED ($n = 93$), 24.8% indicated they had some college, but no degree ($n = 163$), 17.8% had an associates or technical degree ($n = 117$), 25.4% obtained a bachelor's degree ($n = 167$), and 17.0% obtained graduate or professional degree ($n = 112$). The demographic composition of the recruited and final samples relative to quota targets is presented in the supplement. The distribution of political orientation was comparable to a nationally representative U.S. sample which found a mean of 5.2 on a similar 11-point scale (Pew Research Center, 2018). Gender and degree attainment closely matched quota targets, however, participants with technical or community college education were overrepresented.

**Procedure**

Participants were informed that the purpose of the current study was "to better understand how people interact with AI chatbots on randomly chosen topics." Before interacting with the AI chatbot, participants were informed that they would first be presented with demographic questions (such as age, education level, race, gender, etc.), and questions about themselves (their personality and political preferences). These questions, which represented the potential moderators in our research design, assessed political orientation, the Openness to Experience personality dimension of HEXACO (Lee & Ashton, 2018), and the Short Form Need for Cognitive Closure Scale (Roets & Van Hiel, 2011).

Next, participants engaged in five rounds of conversation with the AI chatbot. To encourage thoughtful responses, participants were not able to copy and paste responses and were required to type at least 10 words in each round. The chatbot was programmed to emphasize either Common Ingroup Identity, Dual Identity, or Separate Identity in the three experimental condition or discuss technology in everyday life in a Control condition.

Participants then completed items related to two types of potential mediators: Social Identity Categorizations (Gaertner & Dovidio, 2000) of Latine immigrants in the United States and Pro-Diversity Beliefs (Kauff et al., 2019). Afterwards, three dependent measures involving prosocial support for Latine immigrants were administered. These measures were adapted versions of the Intergroup Giving and Intergroup Acting in Favor of Refugees (IGIAF) Scale (Hanioti et al., 2024) and an Intergroup Helping measure (email sign-ups to receive further information about organizations that assist Latine immigrants, including volunteering). At the end of the session, participants were debriefed.

Details of the materials used in the present research are described, in the order in which they were presented to participants, in the subsequent sections on Moderators, AI Chatbot Interactions, Mediators, and Dependent Variables.

The present study was approved by the ethical review board of the University of Oslo (38418715), and exempted from the Norwegian Agency for Shared Services in Education and Research (477896). The IRB of Yale University further reviewed and exempted the research (2000042073). All participants provided informed consent.

## Moderators

### *Political Orientation*

Participants were asked to indicate their political orientation on a scale from 0 (*very liberal*) to 10 (*very conservative*) in terms of social and economic issues. These two scores were highly correlated, $r = .89$, $p < .001$, and thus averaged as preregistered.

### *HEXACO Openness Items*

Sixteen items taken from the HEXACO scale (Lee & Ashton, 2018) were used to measure openness to experience. Each item was rated on a 5-point scale from 1 (*strongly disagree)* to 5 (*strongly agree*). Example items included "I'm interested in learning about the history and politics of other countries," and "I find it boring to discuss philosophy" (reversed item). An average score was created across the items (after reverse-coding of the reversed items), showing satisfactory reliability ($\alpha = .86$).

### *Short Form Need for Cognitive Closure Scale*

The 15-item short form Need for Cognitive Closure (NFC) Scale (Roets & Van Hiel, 2011) was included to measure participants' desire for certainty. Each item was rated on a 6-point scale from 1 (*strongly disagree*) to 6 (*strongly agree*). Because the original factor structure had unsatisfactory fit in a confirmatory factor analysis (see Supplementary Materials), exploratory factor analyses were conducted. Four factors were identified as the best fitting solution (see Supplementary Materials for details). Situational Certainty contained four items (e.g., "I don't like situations that are uncertain."; α = .87), Routine contained three items (e.g., "I find that establishing a consistent routine enables me to enjoy life more."; α = .90), Problem Resolution contained three items e.g., "When I have made a decision, I feel relieved"; (α = .74), and Consensus contained four items (e.g., "I feel irritated when one person disagrees with what everyone in a group believes"; α = .71). One item from the short form NFC scale did not load onto any factors and was omitted (e.g., "I feel uncomfortable when I don't understand the reason why an event occurred in my life.").

**AI Chatbot Interactions**

Participants were then introduced to the AI chatbot interaction with the following text:

> Conversations between people and AI are taking place every day. This study seeks to understand how humans communicate with chatbots on various topics. In the next part of the study, you will be asked to discuss a randomly chosen topic related to current events with an AI chatbot. Start by telling the AI your viewpoint, and explain why. Then, build on the chatbot's reply to further discuss your thoughts. There will be 5 rounds of conversation. Please avoid sharing personally identifiable information with the chatbot.

Then, participants were randomly assigned to one of four conditions: Control, Separate Identity, Common Ingroup Identity, and Dual Identity. In each condition, participants were presented with the following prompt to initiate conversation with the chatbot:

> The United States has a long history of immigration and continues to evolve demographically. For example, many immigrants come to the U.S. from Latin America. However, people often vary a lot in their opinions about what the best way for these different groups to coexist is. In a sentence or two, try to explain your perspective on this issue.

This introductory message was presented for all groups including the control group to prevent that priming of the topic in the experimental conditions and not in the control condition would confound observed effects.

### *Prompt Engineering and Model Selection*

GPT-4o (OpenAI) was selected as the large language model for the present study based on four considerations. First, the model offered a favorable price relative to comparable models available at the time of data collection, which made a design involving five rounds of individually tailored conversation with a large national sample economically feasible. Second, GPT-4o has strong linguistic ability, producing fluent, coherent, and contextually appropriate language. This capability was essential because the manipulation depended on the chatbot formulating persuasive arguments that were responsive to each participant's specific concerns rather than delivering scripted text. Third, the model's low response latency allowed the dialogue to unfold at a natural conversational pace, reducing the risk that long waiting times between rounds would lead to disengagement or attrition. Finally, and decisively, we were able to access a locally installed version of the model. Because all conversations were processed on infrastructure controlled by the research team, participant data were not shared with any third party and could not be used by other parties for model training, ensuring participant safety and compliance with institutional data protection requirements.

The condition prompts were developed following general prompt engineering principles. Rather than relying on a broad instruction to persuade, each system prompt specified the chatbot's role, the persuasive objective of the assigned condition, the concrete argumentative techniques to be used, and the structure of the interaction. To ensure that any differences between conditions could be attributed to the identity framing itself, the prompts for all conditions were engineered in parallel with an identical structure, length, and level of detail, as described below. The full verbatim prompts are provided in the Supplementary Materials.

### ***General Chatbot Infrastructure and Instructions***

Prompt instructions to the chatbot were housed in a separate Cloudflare server, to ensure they were hidden from participants. Overarchingly, the AI chatbot was instructed that it was participating in a research study about social identity recategorization through persuasive conversation. Its directions related to the conversation structure were to assess the participants' stance, persuade within the directive, reinforce shifts, and monitor the participant's attitude. The chatbot was instructed to end responses with open-ended questions to encourage further dialogue or ask participants for their thoughts on the points raised in rounds 1-4, while round 5 ended with a reflective and definitive statement. Each chatbot response was set to approximately 150 words.

To ensure comparability, each of the prompts used in the following conditions adhered to the same structure and level of detail. All conditions followed the same conversation structure (five rounds of conversation with chatbot responses of approximately 150 words), the engagement style was kept consistent across conditions and focused on argument-based persuasion instead of emotional appeals, and all conditions had the same approach for redirecting off-topic responses back to the topic. The three experimental conditions all focused on White non-Hispanic Americans (participant ingroup) and Latine immigrants (outgroup), differing only in the identity framework being promoted (Common Ingroup Identity, Dual Identity, Separate Identity, or Control). To maintain engagement, in all conditions, the chatbot was instructed to provide new arguments and examples in each round of conversation rather than repeating points that were previously raised. The Control condition differed in terms of the conversational topic (technology in daily life rather than identity) but kept the same structural features as the experimental conditions. For the exact verbatim content of each prompt, please see the Supplementary Materials.

**Common Ingroup Identity condition.** In this condition, the AI chatbot was directed to increase a non-Hispanic White participant's sense of common identity with immigrants from Latin America by emphasizing that all are part of a single shared in-group as residents of the U.S. Core techniques for this condition included emphasizing the shared place, shared challenges, shared future, and shared contributions between immigrant groups and participants, and using inclusive language such as "together" and "our society". The chatbot was also instructed to highlight potential benefits of a common identity, including reduced tension, increased collaboration, and belonging while downplaying differences in favor of common identity categorization.

**Dual Identity condition.** In the Dual Identity condition, the AI chatbot was directed to persuade non-Hispanic White participants that dual identities (e.g., American and Latin American) are valuable and can coexist within a shared American society. Here, the immigrant groups highlighted are the same as the Common Ingroup Identity condition, with additional emphasis on both the U.S. American and immigrant identities being regarded as equally important and compatible. Specific instructions included validating both identities by honoring uniqueness and togetherness, reframing differences as assets that enrich the community, emphasizing values that transcend cultural differences, and highlighting the benefits of dual identity categorization such as reduced tension, collaboration, productivity, innovation and belonging.

**Separate Identity condition.** In the Separate Identity condition, the chatbot was instructed to increase a view of White non-Hispanic Americans and immigrants from Latin

America as separate groups with separate identities, by emphasizing the distinctions and potential incompatibilities between both groups. Instructions included focusing on language barriers, cultural value differences, and why it is meaningful to view these two groups as being separate. Core techniques for this condition included highlighting contrasting social norms and values as well as potential conflicts arising from these differences such as increased tension, weakened collaboration and productivity, and less belonging in American society. The chatbot was also instructed to focus on challenges of coexistence rather than benefits.

**Control Condition.** In the Control condition, participants first responded to the opening prompt regarding immigration and coexistence as in the other conditions. Then, instead of responding directly, the AI chatbot informed participants that the topic was switching, and they would then be discussing the role of technology in their everyday life. The chatbot was instructed to invite participants to reflect on the role of technology in their everyday life, including how their technology use has evolved over time. Additionally, the chatbot was instructed not to mention cultural, ethnic, national, or group identities from this point forward and keep focus strictly on technology use in the U.S., redirecting the conversation back to technology use if necessary.

## Mediators

### *Social Identity Categorizations*

To evaluate the effectiveness of conversation with the AI chatbot on recategorization of immigrants, participants completed a measure containing nine items adapted from Kunst et al. (2015) after interacting with the chatbot. Participants responded to items on a scale of 1 (*strongly disagree*) to 5 (*strongly agree*), with three items measuring one-group categorization (e.g., "For me, immigrants from Latin America are Americans like all of us. I do not see them as members of specific immigrant groups."; α = .87), three items measuring dual categorization (e.g., "For me, immigrants from Latin America are Americans like all of us, but also members of specific immigrant groups."; α = .86), and three items measuring separate categorization (e.g., "I see immigrants from Latin America as a separate group from Americans."; α = .91)

***Pro-Diversity Beliefs Scale***

Pro-diversity beliefs were measured using the Pro-Diversity Beliefs Scale (Kauff et al., 2019) rated from 1 (*completely disagree*) to 5 (*completely agree*). This scale included items such as "A society that is diverse functions better than one that is not diverse," and "I value cultural diversity in the U.S. because it benefits the country." An average score was calculated based on all responses (α = .93).

**Dependent Variables**

***Intergroup Giving and Intergroup Acting in Favor of Refugees Scale***

To determine intentions to engage in prosocial behaviors towards immigrants, participants responded to all 13 items adapted from the Intergroup Giving and Intergroup Acting in Favor of Refugees (IGIAF) Scale (Hanioti et al., 2024). As opposed to referencing refugees specifically as in the original scale, participants were asked how likely they were to engage in behaviors to help immigrants from Latin America within the next six months on a scale of 1 (*Very unlikely*) to 7 (*Very likely*). Six items were designed to measure intentions to give (e.g., "Donating money, food, clothes, toys, etc."; α = .92), and seven items were intended to measure intention to act in favor of immigrants in the U.S. (e.g., "Petitioning to show political discontent"; α = .92).

***Intergroup Helping Behavior***

Participants were then presented with information about three organizations currently helping immigrants in the U.S. This included main initiatives and services provided by the organizations that help immigrants directly. The organizations included Nationalities Service Center (which provides free legal services, English language courses, employment services, and basic needs to immigrants), CIANA (which offers civics classes, a homework help program for children, assistance with applications for social services, and immigration legal services), and Immigrant Welcome Center (which offers a multilingual helpline, provides citizenship classes, and teaches English literacy to immigrants). Each organization was presented on a separate page and in random order. On each page, participants were given the option to provide an email to receive updates from the organization or skip the entry field and continue with the survey. The actual email entries were not recorded for privacy reasons and automatically replaced with a numeric response after they had been verified to be in a standard email format (0 = no email entered; 1 = email entered). An average score was calculated based on all three responses (KR-20 = .85).

***Deviation from Pre-registration: Attention Checks***

A total of three attention checks were presented to participants. One item (“Please select ‘Disagree’ to indicate that you are paying attention”) was randomly presented within the HEXACO openness measure, and another (“Please select ‘Likely’ to indicate that you are paying attention”) was randomly presented within the adapted IGIAF scale. Zero participants failed both attention checks, thus none were excluded based on this pre-registered criterion.

After conversing with the chatbot, participants were also asked what the AI chatbot mostly talked with them about, with the options aligning with each experimental condition: (a) “Talked PRIMARILY about technology usage in everyday life.”, (b) “Talked PRIMARILY about how we need to emphasize our shared common identity as Americans INSTEAD OF focusing on cultural differences.”, (c) “Talked PRIMARILY about how dual identities (e.g., American + country of origin) are valuable and can coexist within a shared American society.”, and (d) “Talked PRIMARILY about differences between immigrants from Latin America and people living in the U.S. and how this creates challenges.” This constituted our third attention check. We had pre-registered that participants who failed this check (regardless of the other two attention checks), would be excluded from analyses. Fail rates were 25.4% in the Dual Identity condition, 8.2% in the Separate Identity condition, and 2% in the control condition, for which the pre-registered exclusions were applied. However, they were unexpectedly high in the Common Ingroup Identity condition (65.2%). An inspection of the responses showed that this was due to 30.4% selecting that the AI chatbot had talked to them about the benefits of dual identity.

Due to the unexpectedly high fail rate, and to rule out the possibility that the chatbot had deviated from the prompt protocol, two independent coders without knowledge of the participants’ assigned conditions blind classified each chatbot conversation into one of the four experimental conditions above. Interrater reliability was 96.8% (Cohen's $\kappa = .958$), and the 26 disagreements (3.2% of conversations) were resolved through discussion. The human coding corresponded closely to participants' actual assigned condition: Control (99.5%), Common Ingroup Identity (97.6%), Dual Identity (93.7%), and Separate Identity (98.0%). These results indicate that the primary reason for participants failing the attention check was not due to the chatbot diverging from explicit instructions for each condition. Rather, the common ingroup identity framing emphasizing shared American identity may have been more difficult to distinguish for participants, as it shares this emphasis on coexistence with the Dual Identity condition.

Having validated that the conversations indeed mapped onto the conditions participants had been assigned to and to ensure statistical power, we loosened the exclusion criterion for the Common Ingroup Identity condition, so that participants who indicated either dual identity or common identity were retained in the final analyses, resulting in a failed attention check rate of 34.8% which was comparable to the Dual Identity condition. We report robustness analyses with the original, stricter exclusion criteria at the end of the Results section and return to the discussion of the potential interventional imprecision of this condition in the Discussion section.

**Analyses**

R version 4.4.0 (R Core Team, 2024) was used for analyses in conjunction with the following packages: tidyverse (v2.0.0; Wickham et al., 2019), psych (v2.6.1; Revelle, 2026), lavaan (v0.6.21; Rosseel et al., 2025; Rosseel, 2012), GPArotation (v3.1; Bernaards & Jennrich, 2005), car (v3.1.2; Fox & Weisberg, 2019), emmeans (v2.0.1; Lenth & Piaskowski, 2025), effectsize (v1.0.1; Ben-Shachar et al., 2020), sjPlot (v2.9.0; Lüdecke, 2025), ggpubr (v0.6.3; Kassambara, 2026), interactions (v1.2.0; Long, 2024), sandwich (v3.1.1; Zeileis et al., 2020), semTools (v0.5-8; Jorgensen et al., 2026), and writexl (v1.5.4; Ooms, 2025). To test H1, H2, and H3, one-way ANOVAs were conducted for each dependent variable with the condition as the independent variable. Planned contrasts compared each experimental condition to the control group. To test H4, a path analysis was implemented using Structure Equation Modeling (SEM) with 5,000 bootstrap iterations for indirect associations. H5, H6, and H7 were tested by a single ordinary least squares multiple regression model. The role of political orientation was tested both linearly and quadratically (to test whether different leaning or general extremity moderated effects). All moderators were mean centered.

## Results

**Descriptive Statistics**

Descriptive statistics for all scale variables are reported in Table 1. Notably, Email Sign-ups, representing a behavioral measure of willingness of participants to share their email address to be contacted by organizations (up to 3) that offer support to immigrants, showed a highly skewed distribution (skewness = 2.89, kurtosis = 7.14), reflecting a generally low level of willingness to receive further information about assistance for immigrants. Correlations between variables are described in Table 2. For example, despite the skewed distribution, email sign-ups correlated positively with pro-diversity beliefs, openness to experience, IGIAF giving, and IGIAF acting, as well as negatively with separate-group categorization.

**Table 1**
*Descriptive Statistics for All Scale Variables*

| Scale | Range | *M* | *SD* | Median | Skewness | Kurtosis |
|---|---|---|---|---|---|---|
| HEXACO Openness | 1–5 | 3.59 | 0.65 | 3.63 | −0.35 | −0.16 |
| NFC Situational Certainty | 1–6 | 4.33 | 1.06 | 4.50 | −0.55 | 0.11 |
| NFC Routine | 1–6 | 4.57 | 1.09 | 4.67 | −0.92 | 0.92 |
| NFC Consensus | 1–6 | 3.42 | 1.00 | 3.25 | 0.01 | −0.21 |
| NFC Problem Resolution | 1–6 | 4.18 | 1.01 | 4.33 | −0.54 | 0.23 |
| Pro-Diversity Beliefs | 1–5 | 3.85 | 1.00 | 4.00 | −0.85 | 0.12 |
| Political Orientation | 0–10 | 4.47 | 3.35 | 4.50 | 0.16 | −1.36 |
| IGIAF Giving | 1–7 | 3.12 | 1.57 | 3.00 | 0.26 | −0.94 |
| IGIAF Acting | 1–7 | 3.57 | 1.67 | 3.57 | 0.09 | −1.09 |
| Email Sign-ups | 0–3 | 0.26 | 0.74 | 0.00 | 2.89 | 7.14 |

*Note.* $N = 658$.

**Table 2**

*Correlations Between Variables*

| Variable | 1 | 2 | 3 | 4 | 5 | 6 | 7 | 8 | 9 | 10 | 11 | 12 | 13 |
|---|---|---|---|---|---|---|---|---|---|---|---|---|---|
| 1. One-Group Categorization | — | | | | | | | | | | | | |
| 2. Dual Categorization | 0.18*** | — | | | | | | | | | | | |
| 3. Separate Categorization | −0.59*** | −0.40*** | — | | | | | | | | | | |
| 4. PDBS (Pro-Diversity Beliefs) | 0.45*** | 0.41*** | −0.61*** | — | | | | | | | | | |
| 5. HEXACO Openness | 0.19*** | 0.18*** | −0.27*** | 0.30*** | — | | | | | | | | |
| 6. NFC Situational Certainty | −0.14*** | 0.03 | 0.15*** | −0.14*** | −0.30*** | — | | | | | | | |
| 7. NFC Routine | −0.06 | −0.04 | 0.15*** | −0.12** | −0.19*** | 0.51*** | — | | | | | | |
| 8. NFC Problem Resolution | −0.05 | 0.03 | 0.10** | −0.06 | −0.20*** | 0.52*** | 0.36*** | — | | | | | |
| 9. NFC Consensus | −0.22*** | −0.07 | 0.30*** | −0.27*** | −0.42*** | 0.57*** | 0.32*** | 0.42*** | — | | | | |
| 10. Political Orientation | −0.33*** | −0.31*** | 0.51*** | −0.65*** | −0.32*** | 0.04 | 0.15*** | 0.04 | 0.18*** | — | | | |
| 11. IGIAF Giving | 0.37*** | 0.16*** | −0.31*** | 0.38*** | 0.30*** | −0.24*** | −0.13*** | −0.04 | −0.26*** | −0.25*** | — | | |
| 12. IGIAF Acting | 0.39*** | 0.24*** | −0.44*** | 0.53*** | 0.43*** | −0.22*** | −0.17*** | −0.07 | −0.29*** | −0.48*** | 0.79*** | — | |
| 13. Email Sign-ups | 0.05 | 0.05 | −0.13*** | 0.13*** | 0.19*** | −0.06 | −0.07 | 0.02 | −0.11** | −0.10** | 0.25*** | 0.28*** | — |

*Note.* $N = 658$. $^{**}p < .01$. $^{***}p < .001$.

**Impact of the Conditions on the Mediators and Outcomes (H1, H2, and H3)**

The hypothesized mediators were measures of social categorization and pro-diversity beliefs. One-way analyses of variance (ANOVAs) of the effects across the four identity manipulation conditions – Common Ingroup Identity, Dual Identity, Separate Identity and Control condition – on each of the three categorization measures yielded general support for H1. The omnibus ANOVA for dual categorization was significant (see Table 3), and the preregistered analysis comparing each identity condition to the Control condition via planned contrasts showed that scores were higher in the Dual Identity condition than in the Control condition. The overall ANOVA for the measure of separate categorization was also significant (see Table 3), with the Common Ingroup Identity condition and the Dual Identity condition each showing a significantly lower level than the control condition (see Table 4). For the measure of one-group categorization, against expectations, there was no overall effect across conditions and none of the planned comparisons were significant (see Tables 3 and 4).

The second hypothesized mediator, in addition to social categorization, was pro-diversity beliefs. The ANOVA did yield an overall significant effect across the four conditions (see Table 3). While the pattern of means conformed to the expectation that diversity attitudes would be more positive in the Common Ingroup Identity and Dual Identity conditions than in the Control condition or the Separate Identity condition, pro-diversity beliefs were not significantly higher in either the Common Ingroup Identity condition or the Dual Identity condition compared to the Control condition (our pre-registered analysis; see Table 4).

**Table 3**

*Experimental Effects on Mediators and Dependent Variables*

| Variable | $df_1$ | $df_2$ | $F$ | $p$ | $\eta_p^2$ | 95% CI |
|---|---|---|---|---|---|---|
| Mediators | | | | | | |
| One-Group Categorization | 3 | 654 | 0.37 | .776 | .002 | [.000, .008] |
| Dual Categorization | 3 | 654 | 4.80 | .003 | .022 | [.003, .045] |
| Separate Categorization | 3 | 654 | 4.56 | .004 | .020 | [.002, .043] |
| Pro-Diversity Beliefs | 3 | 654 | 2.96 | .032 | .013 | [.000, .032] |
| Dependent Variables | | | | | | |
| Willingness to Give | 3 | 654 | 0.83 | .476 | .004 | [.000, .014] |
| Willingness to Act | 3 | 654 | 1.70 | .165 | .008 | [.000, .023] |
| Email Sign-ups | 3 | 654 | 1.20 | .310 | .005 | [.000, .018] |

*Note.* $F$-tests are Type III one-way ANOVAs. $\eta_p^2$ = partial eta-squared. 95% CIs are two-sided. $N = 658$.

**Table 4**

*Planned Contrasts for Mediators*

| Subscale | Contrast M [95% CI], M [95% CI] | *b* | 95% CI | *t* | *p* | *d* |
|---|---|---|---|---|---|---|
| One-Group Categorization | Common Identity: *M* = 4.23 [3.97, 4.48] vs. Control: *M* = 4.06 [3.85, 4.27] | 0.17 | [−0.16, 0.50] | 1.01 | .314 | 0.11 |
| | Dual Identity: *M* = 4.14 [3.90, 4.37] vs. Control: *M* = 4.06 [3.85, 4.27] | 0.08 | [−0.24, 0.40] | 0.48 | .634 | 0.05 |
| | Separate Identity: *M* = 4.09 [3.87, 4.31] vs. Control: *M* = 4.06 [3.85, 4.27] | 0.03 | [−0.27, 0.34] | 0.21 | .835 | 0.02 |
| Dual Categorization | Common Identity: *M* = 4.76 [4.52, 5.01] vs. Control: *M* = 4.66 [4.46, 4.86] | 0.10 | [−0.22, 0.42] | 0.62 | .533 | 0.07 |
| | Dual Identity: *M* = 5.23 [5.00, 5.46] vs. Control: *M* = 4.66 [4.46, 4.86] | 0.57 | [0.26, 0.88] | 3.63 | < .0( | 0.39 |
| | Separate Identity: *M* = 4.81 [4.59, 5.02] vs. Control: *M* = 4.66 [4.46, 4.86] | 0.14 | [−0.15, 0.44] | 0.96 | .335 | 0.10 |
| Separate Categorization | Common Identity: *M* = 2.90 [2.63, 3.16] vs. Control: *M* = 3.40 [3.18, 3.62] | −0.50 | [−0.85, −0.16] | −2.86 | .004 | −0.32 |
| | Dual Identity: *M* = 2.94 [2.69, 3.19] vs. Control: *M* = 3.40 [3.18, 3.62] | −0.46 | [−0.79, −0.12] | −2.70 | .007 | −0.29 |
| | Separate Identity: *M* = 3.34 [3.11, 3.57] vs. Control: *M* = 3.40 [3.18, 3.62] | −0.05 | [−0.37, 0.26] | −0.34 | .733 | −0.04 |
| Pro-Diversity Beliefs | Common Identity: *M* = 3.99 [3.82, 4.16], vs. Control: *M* = 3.78 [3.64, 3.92] | 0.21 | [−0.05, 0.48] | 1.90 | .058 | 0.21 |
| | Dual Identity: *M* = 3.97 [3.81, 4.13] vs. Control: *M* = 3.78 [3.64, 3.92] | 0.19 | [−0.02, 0.50] | 1.75 | .081 | 0.19 |
| | Separate Identity: *M* = 3.72 [3.58, 3.87] vs. Control: *M* = 3.78 [3.64, 3.92] | −0.06 | [−0.25, 0.24] | −0.60 | .554 | −0.06 |
| Willingness to Give | Common Identity: *M* = 3.25 [2.99, 3.52], vs. Control: *M* = 3.07 [2.85, 3.29] | 0.18 | [−0.16, 0.53] | 1.03 | .304 | 0.12 |
| | Dual Identity: *M* = 3.20 [2.95, 3.45] vs. Control: *M* = 3.07 [2.85, 3.29] | 0.12 | [−0.21, 0.46] | 0.73 | .465 | 0.08 |
| | Separate Identity: *M* = 3.00 [2.77, 3.23] vs. Control: *M* = 3.07 [2.85, 3.29] | −0.07 | [−0.39, 0.25] | −0.43 | .666 | −0.04 |
| Willingness to Act | Common Identity: *M* = 3.85 [3.57, 4.14] vs. Control: *M* = 3.45 [3.22, 3.68] | 0.40 | [0.03, 0.77] | 2.15 | .032 | 0.24 |
| | Dual Identity: *M* = 3.54 [3.27, 3.80] vs. Control: *M* = 3.45 [3.22, 3.68] | 0.09 | [−0.27, 0.44] | 0.49 | .624 | 0.05 |
| | Separate Identity: *M* = 3.51 [3.26, 3.75] vs. Control: *M* = 3.45 [3.22, 3.68] | 0.06 | [−0.28, 0.40] | 0.34 | .732 | 0.04 |
| Email Sign-ups | Common Identity: *M* = 0.30 [0.18, 0.43] vs. Control: *M* = 0.27 [0.17, 0.37] | 0.03 | [−0.13, 0.20] | 0.41 | .683 | 0.05 |
| | Dual Identity: *M* = 0.31 [0.19, 0.43] vs. Control: *M* = 0.27 [0.17, 0.37] | 0.04 | [−0.12, 0.20] | 0.53 | .598 | 0.06 |
| | Separate Identity: *M* = 0.17 [0.07, 0.28] vs. Control: *M* = 0.27 [0.17, 0.37] | −0.09 | [−0.25, 0.06] | −1.24 | .216 | −0.13 |

*Note. b* = unstandardized regression coefficient. 95% CI = confidence interval on the contrast. *N* = 658.

We also expected that Common Ingroup Identity and Dual Identity conditions would produce greater intergroup helping intentions and behavior relative to the Control condition (H3). However, we did not find direct effects of the manipulations. Each of the overall ANOVAs for Intergroup Giving, Intergroup Acting, and Email sign-ups (our immediate behavioral measure) was nonsignificant (Table 3). Yet, a significant contrast was observed, in which willingness to act was significantly higher in the Common Ingroup Identity conditions than in the Control condition (see Table 4).

An additional exploratory analysis, a 2 × 2 factorial ANOVA (Superordinate Identity: Yes vs. No × Subgroup Identity: Yes vs. No) showed that willingness to act was greater when the chatbot conversation included an emphasis on common identity (the Common Ingroup Identity and Dual Identity conditions) than when it did not (the Separate Identity and Control conditions), *M*s = 3.69 [3.50, 3.89], vs. 3.48 [3.31, 3.65], $F(1, 654) = 4.61$, $p = .032$, $\eta^2 = .004$. No other effects on helping-related outcomes were statistically significant (*ps* > .156). Thus, there was only limited support for H3 concerning the direct effect of the manipulation of the conversation content on helping-related outcomes.

**Mediation Analysis (H4)**

Previous research on the Common Ingroup Identity Model (Gaertner & Dovidio, 2000) demonstrates that stronger shared identity resulting in a one-group categorization or a dual categorization relates to more positive intergroup orientations, whereas stronger separate categorization is associated with more negative orientations. The pattern of zero-order correlations in Table 2 (except for some correlations with the email sign-ups, for which responses were infrequent) are consistent with these findings. To test H4 by considering these effects simultaneously, we created pathway model using structural equation modeling (SEM) to examine direct and indirect pathway to helping-related outcomes. The full model was fitted with all possible paths from predictors to mediators and from mediators to dependent outcomes. As this model is fully saturated, model fit is not reported.

Figure 1 presents the model, except for correlations between measures, which were estimated but not visualized to ensure clear presentation. These can be found in the Supplementary Materials. Both the Dual Identity and Common Ingroup Identity conditions reduced separate categorization, which subsequently was associated with a higher willingness to act. Indirect effects based on 5,000 bootstrap resamples were significant for both the Common Ingroup Identity condition ($b$ = 0.055, 95% BCa CI [0.018, 0.114]) and Dual Identity condition ($b$ = 0.051, 95% BCa CI [0.017, 0.105]), which both were indirectly linked to a higher willingness to act. Additionally, the mediator one-group categorization was positively associated with willingness to give, and willingness to act. Similarly, pro diversity beliefs were positively associated with willingness to give, and act, in favor of Latine immigrants.

**Figure 1**

*Path Model*

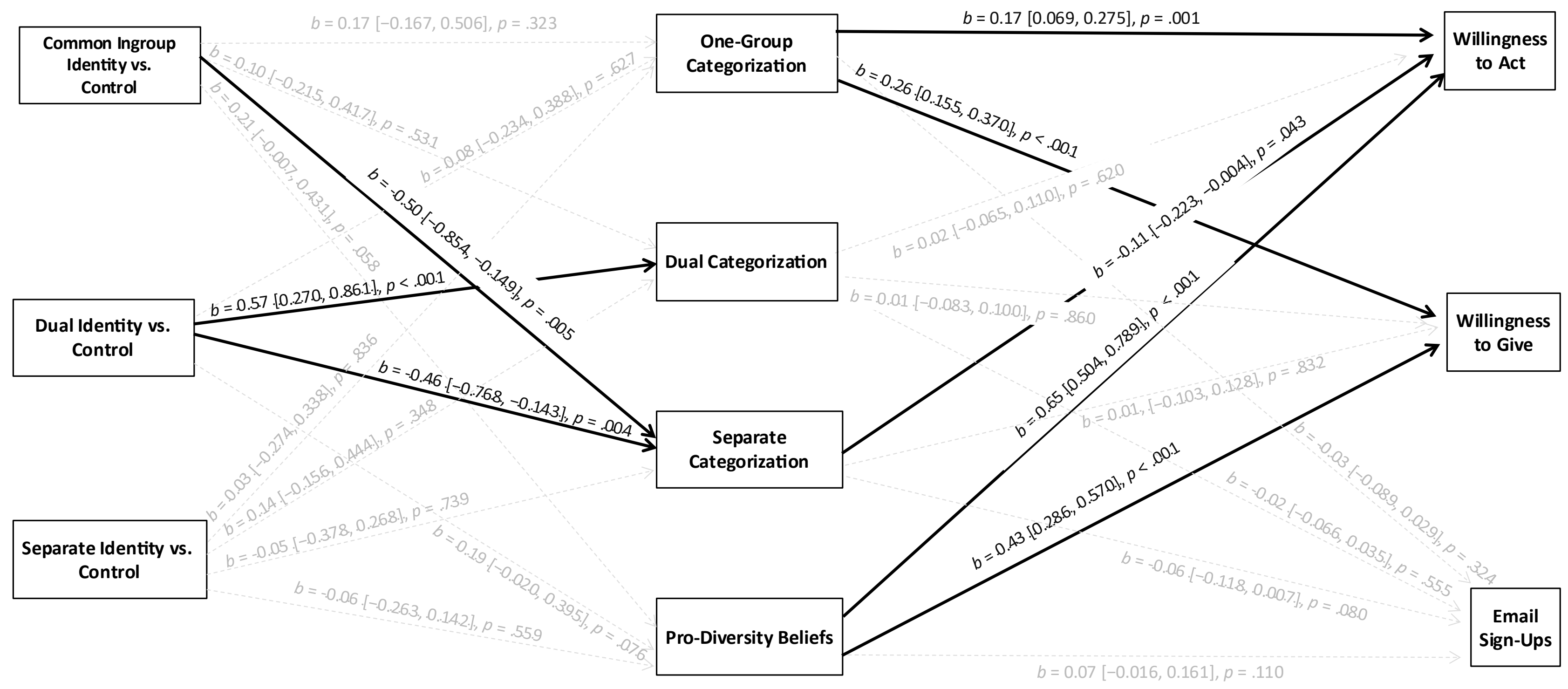

*Note*. Coefficients are unstandardized (*b*). 95% confidence intervals shown. Significant paths ($p < .05$) are depicted in black. Non-significant paths are depicted in gray. $N = 658$

**Moderation Analysis (H5, H6, H7)**

As discussed earlier, we also examined the effects of three types of individual-difference variables – openness to experience, NFC (Situational Certainty, Routine, Problem Resolution, and Consensus), and political orientation in the context of the dynamics described by the Common Ingroup Identity Model. As indicated by the zero-order correlations presented in Table 2 and consistent with previous research on the overall relationship of these variables with intergroup orientations that we reviewed earlier, participants higher on openness to experience, lower on need for closure, and more liberal in their political orientation demonstrated higher levels of giving and acting intentions in Favor of Immigrants. Participants more open to experience and those more politically liberal were also more willing to make themselves available to help with email sign-ups.

We further hypothesized that political orientation, need for cognitive closure, and openness would moderate the effects of experimental conditions on mediators and dependent variables (H5, H6, and H7, respectively). One model was used to test for effects of all moderators for each mediator and outcome variable (see Table S1 and S2 in supplement). To account for the large number of simultaneous tests, Holm-Bonferroni correction was applied to all interaction terms in the models. The only statistically significant interaction that remained was between NFC Situational Certainty and the Dual Identity condition, $b = 0.88$, 95% CI [0.49, 1.27], $p = .001$. As visualized in Figure 2, the Dual Identity condition increased dual categorization among participants scoring high on NFC Situational Certainty but decreased it among those scoring low on it. Thus, whereas the role of NFC obtained some support, the direction of moderation was contrary to our prediction. No other moderation effects survived correction.

**Figure 2**

*Floodlight Plot: Moderating Effect of NFC Situational Certainty on Dual Categorization*

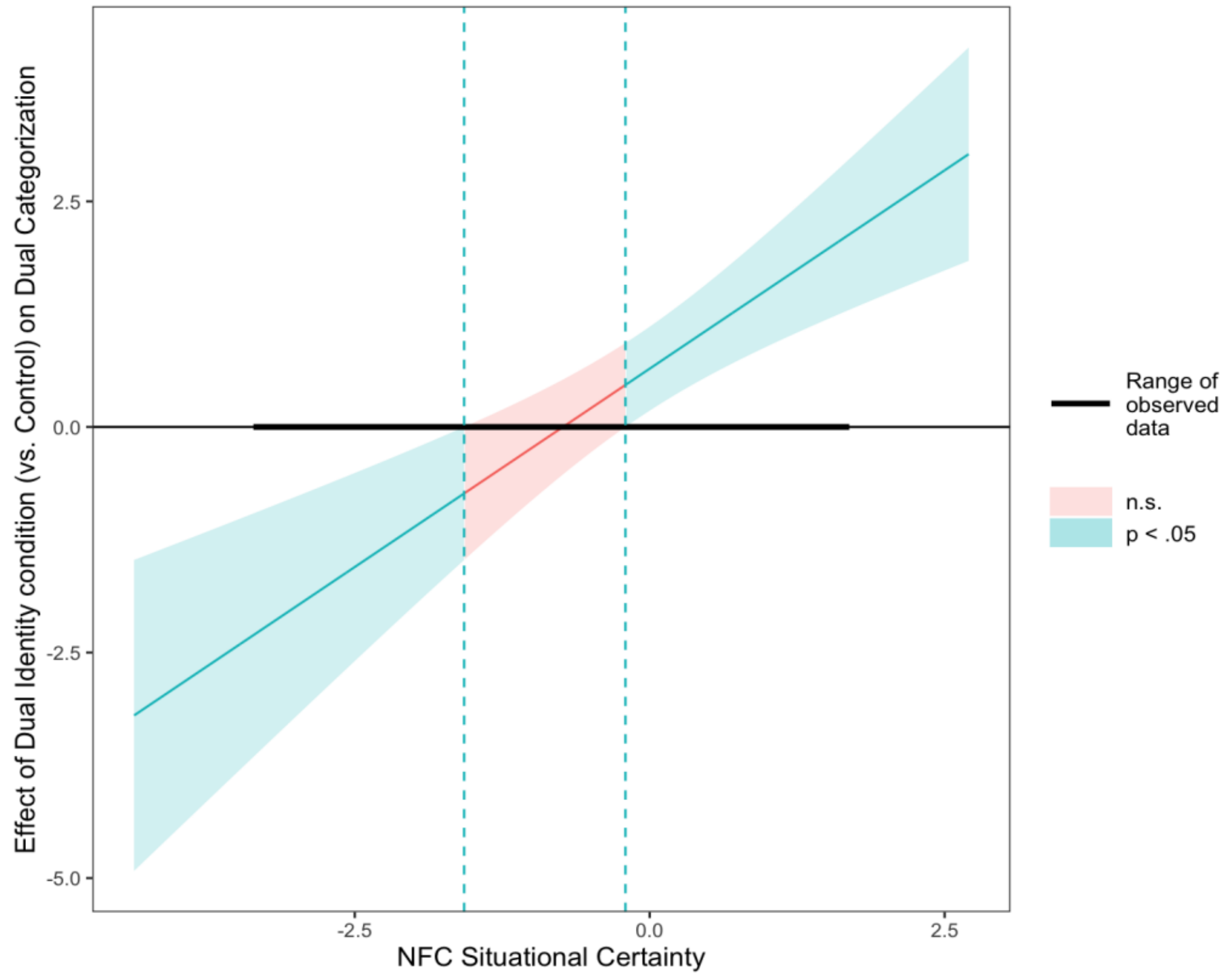

*Note.* The moderator is standardized. Ribbons represent 95% CIs.

## Robustness Check: Analyses with the Stricter Exclusion Criterion

Due to aforementioned deviations from pre-registered criteria in the Common Ingroup Identity condition, we ran strict analyses excluding all participants who failed to identify the correct condition in the attention check ($N = 597$, vs. $N = 658$ originally). The correlational structure, the pattern of significance across the omnibus ANOVAs, and the single surviving moderation effect were unchanged except for minor changes in estimates. Three findings differed: the planned contrast of common identity framing vs control on pro-diversity beliefs reached significance in the stricter sample, as did the path from pro-diversity beliefs to email sign-ups, while the path from separate categorization to willingness to act became non-significant. Full comparison statistics and tables are reported in the Supplementary Materials.

**Exploratory Semantic Similarity Analysis of Conversational Content to the Target Narratives**

For each of the five conversation rounds, we extracted the chatbot's statement and the participant's statement for all participants in the analytic sample ($N = 658$). Each condition was associated with a target narrative that the chatbot was instructed to advance (common ingroup identity, CII; dual identity, DI; separate identity, SI; and a technology-use topic in the control condition). Every statement and every target narrative was embedded with OpenAI's text-embedding-3-large model (3,072 dimensions), and semantic similarity was computed as the cosine similarity between a statement's and a narrative's embeddings. Participant-level scores were the mean similarity across rounds 1–5, computed separately for chatbot and participant statements and for each narrative. Scores were robust to the choice of representation: similarities computed with a second embedding model (text-embedding-3-small) correlated $r = .98$ with the primary scores, and a purely lexical TF-IDF baseline correlated $r = .63$, as expected for a surface-overlap measure.

We compared the prevalence of each theme across conditions with one-way ANOVAs on these similarity scores, separately for chatbot and participant statements, following up with Tukey HSD pairwise comparisons. We then computed, within each identity condition, the partial correlation between participants' similarity to their own condition's target narrative and each behavioral outcome, controlling for similarity to the other two identity narratives, visualized as added-variable plots. Analyses were conducted in Python 3.14 (pandas, SciPy, statsmodels, scikit-learn).

### ***Theme Prevalence Across Conditions***

Conditions differed strongly in the prevalence of every target theme (Table 5, Figure 3). For chatbot statements, $F(3, 654)$ ranged from 3,636 to 5,024 (all $p < .001$, $\eta^2 = .94–.96$); for participant statements, $F(3, 654)$ ranged from 437 to 1,013 (all $p < .001$, $\eta^2 = .67–.82$). The technology topic was most prevalent in the control condition, and each identity theme was substantially more prevalent in all three identity conditions than in the control condition. Within the identity conditions, Tukey comparisons showed that the Dual Identity and Separate Identity conditions were the (joint) most prevalent on their own narratives, whereas the Common Ingroup Identity theme was elevated to a similar degree across all three identity conditions, indicative of the shared American identity also being invoked in dual-identity and separate-identity conversations (see Table 5).

**Table 5**

*Condition Differences in Theme Prevalence: One-Way ANOVAs and Tukey HSD Comparisons*

| Statements | Theme | $F(3, 654)$ | $p$ | $\eta^2$ | Own condition vs. others |
|---|---|---|---|---|---|
| Chatbot statements | CII narrative | 3,986 | < .001 | .95 | < Control, = DI, = SI |
| Chatbot statements | DI narrative | 3,636 | < .001 | .94 | < Control, < CII, < SI |
| Chatbot statements | SI narrative | 4,685 | < .001 | .96 | < Control, < CII, < DI |
| Chatbot statements | Control topic (technology) | 5,024 | < .001 | .96 | < CII, < DI, < SI |
| Participant statements | CII narrative | 534 | < .001 | .71 | < Control, = DI, = SI |
| Participant statements | DI narrative | 437 | < .001 | .67 | < Control, < CII, = SI |
| Participant statements | SI narrative | 553 | < .001 | .72 | < Control, < CII, < DI |
| Participant statements | Control topic (technology) | 1,013 | < .001 | .82 | < CII, < DI, < SI |

*Note.* Similarity scores are participant-level means (rounds 1–5) of cosine similarities between statements and each theme. “>” and “<” denote Tukey HSD $p < .05$ for the theme's own condition compared with each other condition; “=” denotes a nonsignificant difference.

**Figure 3**

*Prevalence of Each Target Theme by Condition*

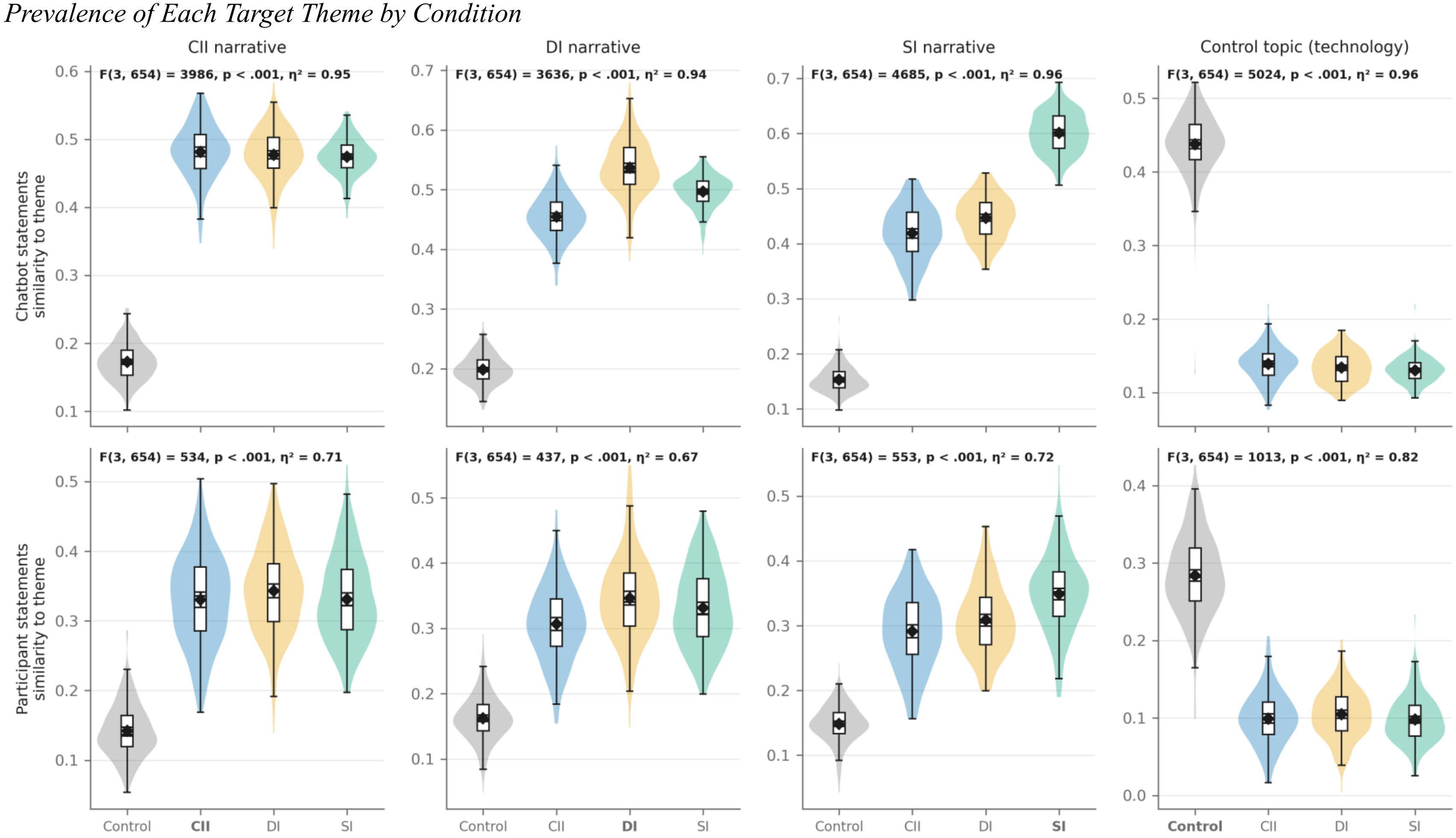

*Note.* Violins show the distributions of participant-level mean similarities (rounds 1–5) between statements and each theme, for chatbot statements (top row) and participant statements (bottom row). Inner boxplots show medians and interquartile ranges; black

diamonds show means with 95% confidence intervals. Panel annotations give one-way ANOVAs across conditions; the theme's own condition is bolded on the horizontal axis.

***Partial Correlations With Behavioral Outcomes***

Within each identity condition, we examined whether participants convergence to their condition's target narrative (controlling for similarity to the other two identity narratives) correlated with different behavioral outcomes (see Table 6, Figure 4). A dissociation emerged. In the CII condition, unique alignment with the shared-identity narrative was positively associated with willingness to act, partial $r = .23$, $p = .009$. In the SI condition, by contrast, unique alignment with the separate-identity narrative was negatively associated with willingness to help, partial $r = -.27$, $p < .001$, and willingness to act, partial $r = -.21$, $p = .005$. Associations in the DI condition were nonsignificant, and email sign-ups showed no positive association in any condition (Table 6). Because narrative similarity was measured rather than manipulated, these associations are correlational process evidence rather than causal effects.

**Table 6**

*Partial Correlations Between Own-Narrative Similarity of Participant Statements and Behavioral Outcomes, by Condition*

| Outcome | Condition | *n* | Partial *r* | 95% CI | *p* |
|---|---|---|---|---|---|
| Email sign-ups (0–3) | Common Ingroup Identity (CII) | 132 | .03 | [-.14, .20] | .743 |
| Email sign-ups (0–3) | Dual Identity (DI) | 151 | -.17 | [-.32, -.01] | .042 |
| Email sign-ups (0–3) | Separate Identity (SI) | 178 | .01 | [-.14, .15] | .942 |
| Willingness to help | Common Ingroup Identity (CII) | 132 | .13 | [-.04, .30] | .132 |
| Willingness to help | Dual Identity (DI) | 151 | .00 | [-.16, .16] | .997 |
| Willingness to help | Separate Identity (SI) | 178 | -.27 | [-.40, -.12] | < .001 |
| Willingness to act | Common Ingroup Identity (CII) | 132 | .23 | [.06, .39] | .009 |
| Willingness to act | Dual Identity (DI) | 151 | -.10 | [-.26, .06] | .207 |
| Willingness to act | Separate Identity (SI) | 178 | -.21 | [-.35, -.06] | .005 |

*Note.* Partial correlations control for similarity to the other two identity narratives. Confidence intervals are based on the Fisher *r*-to-*z* transformation.

**Figure 4**

*Added-Variable Plots of Own-Narrative Similarity and Behavioral Outcomes*

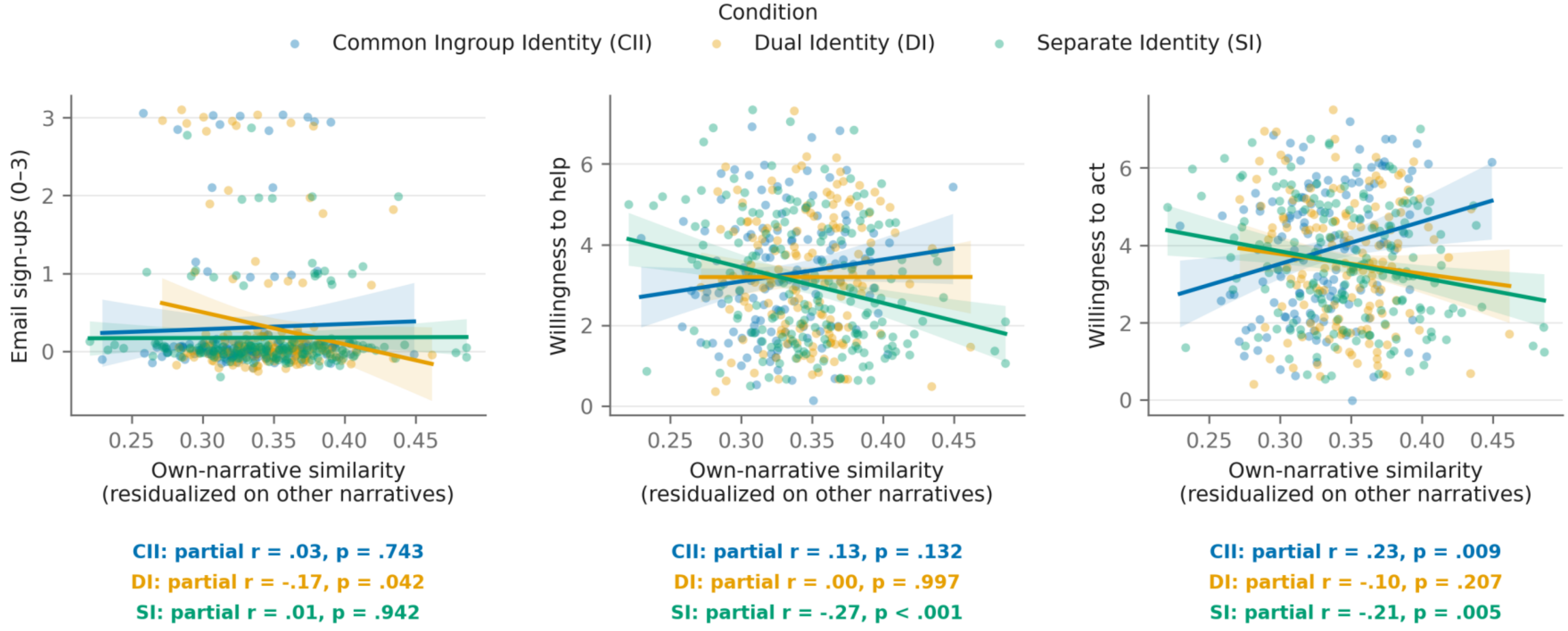

*Note.* Within each condition, both variables are residualized on similarity to the other two identity narratives (condition means added back for readability), so each line depicts the partial association. Points are participants (email sign-ups vertically jittered for visibility); lines are ordinary least squares fits with 95% confidence ribbons; annotations give the partial correlations reported in Table 6.

## Discussion

The present research had two main objectives. One was to apply AI chatbot technology as a research tool to influence the way non-Latine White American participants socially categorize Latine immigrants and affect Pro-Diversity Beliefs. We tested, for the first time, to the best of our knowledge, the potential of using brief interactions between participants and an AI chatbot to deliver a scalable recategorization intervention. The other objective was to investigate the direct and indirect pathways from the chatbot manipulation to helping intentions and behavior through the mediators pro-diversity beliefs and social categorization. Based on work on the Common Ingroup Identity Model (Gaertner & Dovidio, 2000; Gaertner et al., 2015) and research on recategorization processes (Dovidio & Kunst, 2026), we hypothesized that programming the AI chatbot to guide brief (five-round) conversations with participants about Latine immigrants (in terms of common, dual, or separate identity with Latine immigrants) would affect how participants perceived relations with Latine immigrants and pro-diversity beliefs, and ultimately influence the degree to which participants responded pro-socially toward Latine immigrants.

With respect to the first objective, participants' chatbot conversations produced several patterns of results for social category identities as intended. In terms of the overall pattern among the three manipulated identity conditions (see Table 4), dual categorization was highest in the Dual Identity condition, and separate categorization highest in the Separate Identity condition (significantly higher than in the Common Ingroup Identity and Dual Identity conditions). The Separate Identity condition, by contrast, did not significantly alter any categorization relative to the Control condition, suggesting that reinforcing "us versus them" framing was harder to induce through a single conversation than dissolving it. These findings revealed, as we hypothesized, that AI chatbot conversations can shape how people perceive members of another group, which can have significant consequences for intergroup relations (Dovidio & Kunst, 2026; Gaertner & Dovidio, 2000; Gaertner et al., 2015; Kunst & Dovidio, 2026).

Several of our categorization predictions, however, were not supported. We had expected Common Ingroup Identity conversations to raise one-group categorization, and we considered that Dual Identity conversations might do the same, given the conceptual overlap between holding a dual identity and acknowledging a shared superordinate group. Neither expectation was met: One-group categorization did not vary across conditions, and none of the planned contrasts approached significance. We had also predicted that Separate Identity conversations would strengthen separate categorization while weakening dual and one-group categorization, yet this condition shifted none of the three measures relative to control.

Two explanations seem plausible. First, adopting a single, fully merged identity asks more of participants than accepting a dual one. Seeing Latine immigrants as both a distinct group and part of a common American whole is arguably a smaller conceptual step than treating the two groups as genuinely one, and it is therefore more open to movement within a brief, five-round exchange. Coming to view another group as simply part of “us” may require more sustained or repeated contact than our intervention offered. Second, the null effect of the Separate Identity condition fits the broader asymmetry already noted: a single conversation appears better suited to loosening group boundaries than to building them. Participants may discount or resist framing that casts an outgroup in openly adversarial terms, and whatever sense of separateness they already held was unlikely to grow from one short prompt. In sum, the intervention was more effective at fostering a dual identity than a fully common one, and more effective at reducing perceived separateness than at creating it.

While pro-diversity beliefs were somewhat greater in the Common Ingroup Identity condition and the Dual Identity condition than in the control condition, these effects were not statistically significant. One interpretation is that pro-diversity beliefs reflect a more stable, ideologically anchored construct than social categorization, and so are less responsive to a single brief conversation. This is consistent with the strong negative correlation we observed between political orientation and pro-diversity beliefs.

With respect to our second objective, the preregistered tests of direct effects of the AI chatbot manipulation of the two helping intention measures (willingness to give, willingness to act) and the behavioral measure (email sign-ups) were mostly non-significant (except for one significant planned contrast showing that the Common Ingroup Identity condition increased willingness to act and the two-factorial analyses showing that conditions emphasizing a superordinate identity had an effect). However, the pattern of findings was partly consistent with the indirect processes hypothesized in the Common Ingroup Identity Model. As indicated in Table 2, higher levels of one-group categorization and dual categorization were correlated with greater willingness to give and to act, and greater separate categorization was related to less willingness to give and act and lower levels of email sign-ups. These findings are consistent with an established body of literature documenting how a stronger sense of shared identity, either as a one-group categorization or a dual categorization, promotes prosocial intergroup intentions and actions (Gaertner et al., 2015; Levine et al., 2005). Also, stronger pro-diversity beliefs correlated with greater willingness to give and to act, as well as with more email sign-ups. In our path model that tested effects simultaneously, the main indirect path from the manipulations (in terms of each identity condition vs. the Control condition) was from the Common Ingroup Identity condition (vs. Control condition) and Dual Identity condition (vs. Control condition) to lower separate categorization to willingness to act.

The three individual-difference measures that we considered in the current research – political orientation, openness to experience, and need for closure – systematically correlated with the ways participants socially categorized Latine immigrants and with the prosocial measures in the ways expected based on previous research. For instance, people who were more politically liberal and those who were more open to experience were more willing to give and to act in favor of Latine immigrants and had higher levels of email sign-ups; participants lower in NFC were more willing to give and to act. However, only one interaction survived correction: NFC Situational Certainty moderated the effect of the Dual Identity condition, such that participants high on this facet showed stronger dual categorization after the dual identity conversation, whereas those low on it showed weaker dual categorization. This lends H6 partial support, since we had singled out the Dual Identity condition as the one whose reception would hinge on need for closure. The direction, however, is the reverse of what we predicted. We had reasoned that holding two group memberships at once is cognitively ambiguous, and that individuals high in need for closure would resist the dual frame for that reason. Instead, those least comfortable with situational uncertainty were the ones who took it up most readily.

A plausible reading turns on what the Situational Certainty facet captures, namely discomfort with unresolved, novel situations and a pull toward a clear answer when one is available. A guided chatbot conversation is precisely such a situation, and the Dual Identity condition offered a definite, repeatedly stated resolution to it: Latine immigrants can be regarded as a distinct group and as part of a shared American whole. For someone who wants the ambiguity of the moment settled, a firm frame delivered over the course of the conversation may have been something to seize on rather than to reject, in line with the seizing-and-freezing account of need for closure (Webster & Kruglanski, 1994). From this perspective, the conceptual complexity of a dual identity mattered less than the clarity with which the chatbot advanced it. Because the effect emerged for one facet out of four and rested on a single corrected test, we treat it as suggestive and in need of replication rather than as a settled result.

The remaining measures – political orientation, openness to experience, and the other need-for-closure facets – did not moderate the manipulation, contrary to H5, H7, and the broader form of H6. One reading of this is encouraging for application. The categorization shifts we obtained appeared to operate fairly evenly across the people we tested, which is a desirable property for an intervention meant to scale. We would temper that conclusion, though. The main effects were themselves modest and few in number, and the interaction terms were held to a conservative Holm-Bonferroni threshold across many tests, so the general absence of moderation may partly reflect limited power to detect it rather than true uniformity.

While the chatbot manipulation partly produced the pattern of different types of social identities that we hypothesized, and these social identities generally correlated with our prosocial outcomes in the expected ways, the main limitation for our results was that the chatbot manipulation mostly did not, with few exceptions, directly predict significant differences in prosocial outcomes. The absence of reliable impact of our chatbot intervention may have occurred for at least two reasons, one involving the nature of the intervention approach and the other relating to the nature of the intergroup context that we examined.

With respect to the nature of the intervention, the absence of condition effects on behavioral intentions (willingness to give/act) and behavioral action (email sign-ups) is consistent with emerging evidence across two large studies by Hackenburg et al. (2026), which indicated that AI persuasion effects on attitudes may operate though different mechanisms than effects on behavior. Hackenburg et al. found that when using a chatbot to motivate political action, providing argument-based persuasion drove attitude change, but this did not correlate with behavioral change (including signing a petition). Analogously, our findings suggest that chatbot conversation interventions may be well-suited for shifting cognitive categorizations, but insufficient for directly motivating behavioral action, which likely requires separate and specific strategies such as commitment escalation (Hackenburg et al., 2026).

While we note that traditional diversity and anti-bias interventions, often much longer in duration, also often fail to improve intergroup behaviors (Devine & Ash, 2026), another reason why observed limited direct effects of our chatbot intervention may involve the intergroup context that we studied – orientations toward Latine immigrants. This context might have posed particular challenges for changing intentions or behavior. First, it is important to acknowledge that how people think, feel, and act toward immigrants is grounded in multiple factors, such as realistic threat, perceived competition, and symbolic threats about undermining traditional values (Esses & Sutter 2026; Kukharkin et al., 2026). How people view the extent of common ingroup, dual, or separate identities is just one of many influences. Second, we note that we aimed to change intentions and behaviors toward a group that has been traditionally negatively stereotyped and for which current events have created very polarized views in the United States. Latine immigrants are associated with longstanding stereotypes that characterize them as low-skilled and criminal. In addition, Latine immigration is often by people associated with undocumented immigration, which generally evokes negative reactions across the political spectrum. This was reflected in the current study, as concerns regarding legal status were raised explicitly by participants, often in the first round of conversation and regardless of condition (see

Supplementary Materials). In addition, intense political rhetoric against Latine immigration, which was occurring at the time we were conducting our experiment, and strong political polarization around immigration policy likely further crystallized orientations toward Latine immigrants, reducing the opportunity and ability to reduce behavior-related change.

Nevertheless, the indirect and few direct effects of the chatbot manipulation suggest some opportunity for positive change. For instance, previous research on the Common Ingroup Identity Model has shown that increasing one-group and dual categorization and decreasing separate categorization reduces fear, increases empathy, and increases willingness to engage in intergroup contact – all of which contribute to subsequent reductions in intergroup bias (Gaertner et al., 1996; Kunst & Dovidio, 2026). Thus, while the chatbot manipulations, which occurred as part of a five round conversation, may not have been potent enough to produce a direct, immediate change in helping intentions and behavior in support of Latine immigrants, it may help foster additional orientations that may produce the kinds of subsequent experiences (e.g., positive intergroup contact) that further motivate stronger prosocial orientation to develop over time. Creating a stronger sense of common ingroup identity, in terms of either a one-group or a dual identity, can increase willingness to engage in positive intergroup contact (Čehajić-Clancy et al., 2023; Yetkili et al., 2025), which in turn is a powerful factor for producing positive intergroup relations even among groups initially viewed in competitive or threatening terms.

The findings of our path model (Figure 1), which simultaneously considers the relations of common ingroup, dual, and separate categorization (along with a pathway involving pro-diversity beliefs) with prosocial intentions and behavior, highlights an additional issue. While the pattern of zero-order correlations aligns closely with expectations derived from the Common Ingroup Identity Model, in the current context separate categorization played a more influential role than either one-group or dual categorization – both in terms of the significant impact of the Common Ingroup Identity condition and Dual Identity condition reducing separate categorization (relative to the control condition) and how weaking separate categorization then was the only statistically significant social categorization path to greater willingness to act and to give in favor of Latine immigrants (but note that the mediation was not significant when applying the stricter exclusion criteria that resulted in lower statistical power). These results suggest that for highly politically and socially charged issues, such as Latine immigration, interventions might initially prioritize reducing separate categorization to alleviate intergroup tension and competitiveness, which can occur when interventions emphasize common identity or dual identity (Kunst & Dovidio, 2026). For instance, Kelman and his colleagues conducted workshops over a several years to improve Palestinian-Israeli relationships in the Middle East (see Kelman, 2018). These researchers introduced elements emphasizing commonality at various stages of the workshops, first undermining perceptions of relations between the groups as zero-sum, in which one group's gains necessarily mean the other group's loss, and then later in the context of interdependent actions for achieving common goals. Kelman (1999) emphasized that an important initial step in his intervention to reduce conflict between Israelis and Palestinians is to reframe the longstanding negative interdependence between these groups in terms of "most notably the positive interdependence between the two groups that exists in reality" (p. 581).

In our research, the separate identity pathway also operated independently of pro-diversity beliefs, although pro-diversity beliefs themselves were strongly associated with willingness to give and act. Together these results suggest dual routes (a cognitive recategorization route operating through reduced separation, and an attitudinal route operating through pro-diversity beliefs) that may be targeted separately or jointly in future interventions.

The exploratory semantic similarity analyses of the conversational content add process evidence to these conclusions. At the level of what was actually said, the manipulations operated as intended: each condition's target narrative was most prevalent in that condition's conversations, not only in the chatbot's statements but also in participants' own statements. That participants themselves came to speak in terms of the assigned narrative indicates that the manipulation was not merely delivered but actively taken up in dialogue, a form of verification that self-report manipulation checks cannot provide. One pattern in these analyses is also informative about the asymmetries discussed above. The common ingroup identity theme was elevated to a similar degree across all three identity conditions, indicating that a shared American identity surfaced even in conversations designed to emphasize dual or separate identities. This spillover may partly explain why the Separate Identity condition failed to strengthen separate categorization: even conversations intended to sharpen group boundaries appear to have carried superordinate identity content.

The correlations between participants' narrative uptake and the outcome measures complement the path model in a similar way. In the Common Ingroup Identity condition, participants whose statements uniquely converged with the shared-identity narrative reported greater willingness to act, whereas in the Separate Identity condition, unique convergence with the separate-identity narrative was associated with lower willingness to help and to act. This dissociation mirrors, at the level of conversational content, the central role of separate categorization in the path model: language that loosened group boundaries tracked stronger prosocial intentions, and language that reinforced them tracked weaker ones. Notably, email sign-ups showed no positive association with narrative uptake in any condition, consistent with the intention-behavior gap discussed above. Because narrative convergence was measured rather than manipulated, these associations are correlational, and participants already favorably disposed toward Latine immigrants may simply have been more willing to echo shared-identity framing. Even so, the agreement between the experimental effects, the self-reported categorization measures, and the conversational content strengthens the interpretation that dissolving us-versus-them boundaries, rather than affirming diversity as such, was the active ingredient of the intervention.

**Limitations**

Several limitations of this study warrant consideration. First, participant interactions with the chatbot was brief (five rounds) and occurred only once. Responses were assessed only once after the chatbot conversation. Future work might productively consider two types of longitudinal designs. One type of longitudinal design could involve, after the one conversation, periodic reassessments of the mediators and outcomes to investigate whether effects persist, decay, or compound over time. A second type of longitudinal study could include repeated chatbot conversations over time. These repeated conversations would reinforce the manipulations of the various forms of social categorization that we studied, and they would provide opportunities for consolidation of the information presented. Consolidation involves neurobiological and cognitive processes through which newly acquired information becomes more stable, durable, and integrated with existing knowledge over time (Dudai et al., 2015). This process would likely strengthen the impact of the manipulations on social categorization and pro-diversity beliefs, as well as increase the likelihood that the intervention would affect behavioral intentions and behavior.

An additional potential limitation of the current research is that the chatbots were programmed to primarily focus on presenting rational arguments in their conversations to change participants' social categorization. There is also a substantial affective component to intergroup relations generally and arguably particularly with the case of Latine immigration (Smith et al., 2007). Future research might expand chatbot instructions to include emotional appeals (for example, perspective-taking prompts or narrative testimony) to test whether these produce stronger outcomes, either independently or in combination with reason-based arguments.

Another limitation relates to our behavioral outcome measure – email sign-ups. Email sign-ups represent a costlier behavioral commitment than the self-reported intention measures, requiring participants to provide contact information and accept future outreach. It was likely viewed as costlier not only of the personal time and effort involved but also because of a concern that it would increase the likelihood of a broader range of appeals by these and related organizations in the future. As a possible consequence, the email sign-up measure showed a pronounced floor effect, with most participants not signing up for any organization. The resulting restricted variance limits sensitivity to detect both experimental effects and mediator associations. There are at least two directions future research can take to address this limitation. One possibility would be to include additional elements in the conversation to strengthen the links to behavior. For instance, research by Gollwitzer and Sheeran (2006) on implementation intentions shows how individuals can reduce the gap between good intentions and concrete action by reflecting on when, where, and how they will execute the steps to achieving their goal. Translating cognitive recategorization into prosocial action may require more than identity framing alone: pairing AI recategorization with explicit action prompts, lowering perceived behavioral cost, or strengthening perceived efficacy, consistent with the broader intention-behavior gap documented in the prosocial behavior literature (Hackenburg et al., 2026; Sheeran & Webb, 2016). A second possibility to be considered in future research is including several alternative behavioral measures varying in costliness and perhaps in other dimensions, such as acts that may involve types of emotionally supportive assistance (e.g., offering advice or encouragement) or instrumental help (e.g., monetary donations) to better understand the nature of the pathways to helping.

A further limitation concerns the inferential status of the indirect effects we report. Because condition was randomly assigned, the paths running from the manipulation to the proposed mediators reflect a genuine causal effect. The paths from those mediators to the helping outcomes, by contrast, rest on measured associations rather than on manipulation, and are thus correlational. Each indirect effect therefore combines one experimentally established link with one observed link, which leaves the mediator-to-outcome segment open to confounding by unmeasured variables and to reverse causation between the mediator and outcome. Claims that a particular shift in categorization produced greater willingness to help should accordingly be read as provisional. A more stringent test would adopt an experimental-causal-chain design (Spencer et al., 2005), in which the mediator is itself manipulated in a separate study, so that both links in the chain are established causally rather than one being inferred from covariation (Bullock et al., 2010).

We also consider that the majority of participants raised the legal status of Latine immigrants as a limitation. Because the chatbot did not distinguish between documented and undocumented Latine immigrants, instead making arguments outside legal status, conversations about social cohesion may have been less persuasive for participants who exclude undocumented immigrants from their vision of coexistence or consider them to be an additional subgroup within a Latine-American identity that they refuse to share a superordinate identity with.

Lastly, the environmental cost of deploying LLMs as a large-scale intervention in terms of energy and water consumption, and carbon emissions must be weighed against the potential benefits of any intervention. Predicted benefits should be commensurate with this cost.

## Summary and Conclusion

The two main aims of the present study were (a) to demonstrate how AI chatbot technology can be a valuable tool for research and application to produce more positive and supportive intergroup relations, and (b) to apply this tool to relations between non-Latine American participants and Latine immigrants to investigate the direct and indirect pathways from the chatbot social identity manipulation (Common Ingroup Identity, Dual Identity, Separate Identities vs. the Control condition) to helping intentions and behavior in support of Latine immigrants. We found, consistent with work on the Common Ingroup Identity Model (Gaertner & Dovidio, 2000; see also Dovidio & Kunst, 2026), that a chatbot intervention to emphasize a common ingroup identity or a dual identity (involving recognition of common and also distinct group identities) had an indirect relationship, independent of pro-diversity beliefs, with intentions to give and act in support of Latine immigrants. While stronger one-group and dual categorizations both correlated, as expected, with intentions to support Latine immigrants, the main pathway was the reduction of separate categorization.

The findings of this study contribute to understanding of how an AI chatbot can be used to change how people mentally represent ingroups and outgroups and promote prosocial intergroup behaviors. LLM chatbots offer a dynamic and individualized platform through which social recategorization may be achieved, going beyond static interventions. The research's success in shifting categorization extends previous AI chatbot persuasion research (Costello et al., 2024; Hackenburg et al., 2025) into the domain of social identity and intergroup helping.

However, the present findings suggest that further refinement is needed to translate social recategorization into prosocial behavior. Our finding that dual and common identity framings were associated with willingness to act through reducing separate categorization, rather than through pro-diversity beliefs, suggests that recategorization interventions may be most effective when they target the dissolution of "us versus them" boundaries, in addition to or instead of focusing on building positive attitudes toward diversity. This boundary-reduction route may also be more tractable in brief interventions concerning highly charged intergroup issues than attitudinal change, which appears to require longer or more emotionally engaging exposure.

We conclude by acknowledging current controversy and concern about the use of AI generally and with respect to social relations specifically (Grodzinsky et al., 2024; Kunst et al., 2026; Lavie-Driver & van der Linden, 2026; Simchon et al., 2024). As the present research demonstrates, though, AI chatbots may be productively harnessed to attenuate rather than to amplify intergroup divisions and promote polarization. Caution remains warranted, however: Longer-term, repeated, or more sophisticated divisive interventions could plausibly yield different outcomes, and the dual-use potential of persuasive LLMs continues to motivate ethical safeguards in their development and deployment (Jones & Bergen, 2026). Also, deploying LLMs for large-scale interventions comes with substantial environmental costs in terms of consumption of energy and water and increased carbon emissions (Ren et al., 2024). These costs need to be weighed against the potential benefits of any intervention.

**References**

Azoulay, P., Jones, B. F., Kim, J. D., & Miranda, J. (2022). Immigration and entrepreneurship in the United States. *American Economic Review: Insights*, *4*(1), 71–88. https://doi.org/10.1257/aeri.20200588

Banfield, J. C., & Dovidio, J. F. (2013). Whites' perceptions of discrimination against Blacks: The influence of common identity. *Journal of Experimental Social Psychology*, *49*(5), 833–841. https://doi.org/10.1016/j.jesp.2013.04.008

Baranger, D. A. A., Finsaas, M. C., Goldstein, B. L., Vize, C. E., Lynam, D. R., & Olino, T. M. (2023). Tutorial: Power analyses for interaction effects in cross-sectional regressions. *Advances in Methods and Practices in Psychological Science, 6*(3), Article 25152459231187531. https://doi.org/10.1177/25152459231187531

Ben-Shachar, M. S., Lüdecke, D., & Makowski, D. (2020). Effectsize: Estimation of effect size indices and standardized parameters. *Journal of Open Source Software*, *5*(56), 2815. https://doi.org/10.21105/joss.02815

Bernaards, C. A., & Jennrich, R. I. (2005). Gradient projection algorithms and software for arbitrary rotation criteria in factor analysis. *Educational and Psychological Measurement*, *65*(5), 676–696. https://doi.org/10.1177/0013164404272507

Bezrukova, K., Spell, C. S., Perry, J. L., & Jehn, K. A. (2016). A meta-analytical integration of over 40 years of research on diversity training evaluation. *Psychological Bulletin*, *142*(11), 1227–1274. https://doi.org/10.1037/bul0000067

Billig, M., & Tajfel, H. (1973). Social categorization and similarity in intergroup behaviour. *European Journal of Social Psychology*, *3*(1), 27–52. https://doi.org/10.1002/ejsp.2420030103

Brewer, M. B., Buchan, N. R., Ozturk, O. D., & Grimalda, G. (2023). Parochial altruism and political ideology. *Political Psychology*, *44*(2), 383–396. https://doi.org/10.1111/pops.12852

Brown, A. (2022, June 7). About 5% of young adults in the U.S. say their gender is different from their sex assigned at birth. *Pew Research Center*. https://www.pewresearch.org/short-reads/2022/06/07/about-5-of-young-adults-in-the-u-s-say-their-gender-is-different-from-their-sex-assigned-at-birth/

Bullock, J. G., Green, D. P., & Ha, S. E. (2010). Yes, but what's the mechanism? (don't expect an easy answer). *Journal of Personality and Social Psychology*, *98*(4), 550–558. https://doi.org/10.1037/a0018933

Čehajić-Clancy, S., Jankovic, A., Opacin, N., & Bilewicz, M. (2023). The process of becoming 'we' in an intergroup conflict context: How enhancing intergroup moral similarities leads to common-ingroup identity. *British Journal of Social Psychology*, 62(3), 1251–1270. https://doi.org/10.1111/bjso.12632

Charnysh, V., Lucas, C., & Singh, P. (2015). The ties that bind: National identity salience and pro-social behavior toward the ethnic other. *Comparative Political Studies*, *48*(3), 267–300. https://doi.org/10.1177/0010414014543103

Costello, T. H., Pennycook, G., & Rand, D. G. (2024). *Durably reducing conspiracy beliefs through dialogues with AI*. Center for Open Science. https://doi.org/10.31234/osf.io/xcwdn

Devine, P. G., & Ash, T. L. (2026). Diversity training. In V. M. Esses, J. F. Dovidio, J. Jetten, D. Sekaquaptewa, & K. West (Eds.), *Sage handbook of psychological perspectives on*

*diversity, equity, and inclusion* (pp. 283–298). Sage. https://doi.org/10.4135/9781036237455.n23

Dobbin, F., & Kalev, A. (2016). Why diversity programs fail. *Harvard Business Review,* 94(7), 14. https://www.researchgate.net/profile/Alexandra-Kalev/publication/360939244_Why_Diversity_Programs_Fail/links/62a075c0c660ab61f86b61ce/Why-Diversity-Programs-Fail.pdf

Dovidio, J. F., & Kunst, J. R. (2026). From "Us and them" to "We": Promise and pitfalls of common ingroup identity. In V. M. Esses., J. F. Dovidio, J. Jetten, D. Sekaquaptewa, & K. West (Eds.), *Sage handbook of psychological perspectives on diversity, equity, and inclusion* (pp. 221–235). Sage. https://doi.org/10.4135/9781036237455.n19

Dudai, Y., Karni, A., & Born, J. (2015). The consolidation and transformation of memory. *Neuron*, *88*(1), 20–32. https://doi.org/10.1016/j.neuron.2015.09.004

Echterhoff, G., Higgins, E. T., & Groll, S. (2005). Audience-tuning effects on memory: The role of shared reality. *Journal of Personality and Social Psychology*, *89*(3), 257–276. https://doi.org/10.1037/0022-3514.89.3.257

Esses, V. M. (2021). Prejudice and discrimination toward immigrants. *Annual Review of Psychology*, *72*(1), 503–531. https://doi.org/10.1146/annurev-psych-080520-102803

Esses, V. M., & Sutter, A. (2026). Diversity, equity, and inclusion for immigrants. In V. M. Esses, J. F. Dovidio, J. Jetten, D. Sekaquaptewa, & K. West (Eds.), *Sage handbook of psychological perspectives on diversity, equity, and inclusion* (pp. 468-484). Sage. https://doi.org/10.4135/9781036237455.n36

Fox, J., & Weisberg, S. (2019). *An R companion to applied regression* (3rd ed.). Sage.

Gaertner, S. L., & Dovidio, J. F. (2000). *Reducing intergroup bias: The Common Ingroup Identity Model*. Taylor and Francis.

Gaertner, S. L., Dovidio, J. F., & Bachman, B. A. (1996). Revisiting the contact hypothesis: The induction of a common ingroup identity. *International Journal of Intercultural Relations*, *20*(3-4), 271–290. https://doi.org/10.1016/0147-1767(96)00019-3

Gaertner, S. L., Dovidio, J. F., Guerra, R., Hehman, E., & Saguy, T. (2015). A common ingroup identity: Categorization, identity, and intergroup relations. In T. D Nelson (Ed.), *Handbook of prejudice, stereotyping, and discrimination* (pp. 433–454). Psychology Press.

Gollwitzer, P. M., & Sheeran, P. (2006). Implementation intentions and goal achievement: A meta-analysis of effects and processes. *Advances in Experimental Social Psychology, 38,* 69–119. https://doi.org/10.1016/s0065-2601(06)38002-1

Grodzinsky, F. S., Wolf, M. J., & Miller, K. W. (2024). Ethical issues from emerging AI applications: Harms are happening. *Computer*, *57*(2), 44–52. https://doi.org/10.1109/mc.2023.3332850

Hackenburg, K., Ibrahim, L., Tappin, B. M., & Tsakiris, M. (2025). Comparing the persuasiveness of role-playing large language models and human experts on polarized US political issues. *AI & SOCIETY*, 1–11.

Hackenburg, K., Hewitt, L., Wagner, C., Tappin, B. M., & Summerfield, C. (2026). Artificial intelligence can persuade people to take political actions. *arXiv preprint arXiv:2604.09200*.

Hanioti, M., Roblain, A., Azzi, A., & Licata, L. (2024). A helpful measure to measure help: The construction and validation of the Intergroup Giving and Intergroup Acting in Favor of Refugees Scale (IGIAF). *International Review of Social Psychology*, *37*(1), 8. https://doi.org/10.5334/irsp.832

Hanson, K., O'Dwyer, E., & Lyons, E. (2021). The national divide: A social representations approach to US political identity. *European Journal of Social Psychology*, *51*(4–5), 833–846. https://doi.org/10.1002/ejsp.2791

Hatungimana, W. (2024). What has nation building got to do with immigration? *National Identities*, *26*(1), 1–24. https://doi.org/10.1080/14608944.2023.2267477

Hehman, E., Gaertner, S. L., Dovidio, J. F., Mania, E. W., Guerra, R., Wilson, D. C., & Friel, B. M. (2012). Group status drives majority and minority integration preferences. *Psychological Science*, *23*(1), 46–52. https://doi.org/10.1177/0956797611423547

Hickel, F., & Bredbenner, M. (2020). Economic vulnerability and anti-immigrant attitudes: Isolated anomaly or emerging trend? *Social Science Quarterly*, *101*(4), 1345–1358. https://doi.org/10.1111/ssqu.12814

Hilbig, B. E., Glöckner, A., & Zettler, I. (2014). Personality and prosocial behavior: Linking basic traits and social value orientations. *Journal of Personality and Social Psychology*, *107*(3), 529–539. https://doi.org/10.1037/a0036074

Huo, Y. J., Dovidio, J. F., Jiménez, T. R., & Schildkraut, D. J. (2018). Not just a national issue: Effect of state-level reception of immigrants and population changes on intergroup attitudes of Whites, Latinos, and Asians in the United States. *Journal of Social Issues*, *74*(4), 716–736. https://doi.org/10.1111/josi.12295

Jones, C., & Bergen, B. (2026). Lies, damned lies, and language statistics: A comprehensive review of risks from manipulation, persuasion, and deception with large language models. *Artificial Intelligence Review*, *59*(4). https://doi.org/10.1007/s10462-026-11517-6

Jorgensen, T. D., Pornprasertmanit, S., Schoemann, A. M., & Rosseel, Y. (2026). *semTools: Useful tools for structural equation modeling* (R package version 0.5-8). https://CRAN.R-project.org/package=semTools

Jost, J. T., & van der Toorn, J. (2011). System Justification Theory. In E. T. Higgins, P. A. M. Van Lange, & A. W. Kruglanski (Eds.), *Handbook of theories of social psychology* (Vol. 2, pp. 313–343). SAGE. https://doi.org/10.4135/9781446249222.n42

Kassambara, A. (2026). *ggpubr: 'ggplot2' based publication ready plots* (R package version 0.6.3). https://CRAN.R-project.org/package=ggpubr

Kauff, M., Stegmann, S., van Dick, R., Beierlein, C., & Christ, O. (2019). Measuring beliefs in the instrumentality of ethnic diversity: Development and validation of the Pro-Diversity Beliefs Scale (PDBS). *Group Processes & Intergroup Relations*, *22*(4), 494–510. https://doi.org/10.1177/1368430218767025

Kawakami, K., Amodio, D., & Hugenberg, K. (2017). Intergroup perception and cognition. *Advances in Experimental Social Psychology*, 55, 1–80. Elsevier. https://doi.org/10.1016/bs.aesp.2016.10.001

Kelman, H. C. (1999). The interdependence of Israeli and Palestinian national identities: The role of the other in existential conflicts. *Journal of Social Issues*, *55*(3), 581–600. https://doi.org/10.1111/0022-4537.00134

Kelman, H. C. (2018). A one-country/two-state solution to the Israeli–Palestinian conflict (2011) 1. In P. Mattar & N. Caplan (Eds.), *Transforming the Israeli–Palestinian conflict* (pp. 203–218). Routledge. https://doi.org/10.4324/9781315170497-17

Kiehne, E., & Ayón, C. (2016). Friends or foes: The impact of political ideology and immigrant friends on anti-immigrant sentiment. *The Journal of Sociology & Social Welfare*, *43*(3). https://doi.org/10.15453/0191-5096.2867

Krol, R. (2021). Effects of immigration on entrepreneurship and innovation. *The Cato Journal*, *41*(3), 551–569. https://doi.org/10.36009/CJ.41.3.5

Kukharkin, A., Barber, F., Cooley, E., Caluori, N., Brown, X., Singh, A., Cipolli, W., & Brown-Iannuzzi, J. L. (2026). White Americans' feelings of being "last place" are associated with anti-DEI attitudes, Trump support, and Trump vote during the 2024 U.S. presidential election. *advances.in/psychology, 1*, e549398. https://doi.org/10.56296/aip00046

Kunst, J. R., & Dovidio, J.F. (2026). Two identities, one alliance? Catalysts and constraints of dual identification for allyship formation. *Current Directions in Psychological Science.* Online advance article.

Kunst, J. R., Obaidi, M., Gollwitzer, A., Brandtzæg, P. B., Hinrichs, Y., Saini, N., & Schroeder, D. T. (2026). Intelligent systems, vulnerable minds: A framework for radicalization to violence in the age of AI. *Personality and Social Psychology Review*, *30*(3), 395–426. https://doi.org/10.1177/10888683261430089

Kunst, J. R., & Thomsen, L. (2015). Prodigal sons: Dual Abrahamic categorization mediates the detrimental effects of religious fundamentalism on Christian–Muslim relations. *The International Journal for the Psychology of Religion*, *25*(4), 293–306. https://doi.org/10.1080/10508619.2014.937965

Kunst, J. R., Thomsen, L., Sam, D. L., & Berry, J. W. (2015). "We are in this together": Common group identity predicts majority members' active acculturation efforts to integrate immigrants. *Personality and Social Psychology Bulletin*, *41*(10), 1438–1453. https://doi.org/10.1177/0146167215599349

Lavie-Driver, N., & van der Linden, S. (2026). Weaponising the past: An extended SIMCA model for how social identity and collective memory shape variation in collective action responses to democratic backsliding. *Advances.in/Psychology*, *1*(1), e719332. https://doi.org/10.56296/aip00055

Lee, K., & Ashton, M. C. (2018). Psychometric properties of the HEXACO-100. *Assessment*, *25*, 543-556. https://doi.org/10.1177/1073191116659134

Lenth, R. V., & Piaskowski, J. (2025). *Emmeans: Estimated marginal means, aka least-squares means* (R package version 2.0.1). https://CRAN.R-project.org/package=emmeans

Levine, M., Prosser, A., Evans, D., & Reicher, S. (2005). Identity and emergency intervention: How social group membership and inclusiveness of group boundaries shape helping behavior. *Personality & Social Psychology Bulletin, 31*(4), 443–453. https://doi.org/10.1177/0146167204271651

Long, J. A. (2024). *Interactions: Comprehensive, user-friendly toolkit for probing interactions* (R package version 1.2.0). https://doi.org/10.32614/CRAN.package.interactions

Lüdecke, D. (2025). *sjPlot: Data visualization for statistics in social science* (R package version 2.9.0). https://CRAN.R-project.org/package=sjPlot

Ooms, J. (2025). *Writexl: Export data frames to Excel 'xlsx' format* (R package version 1.5.4). https://CRAN.R-project.org/package=writexl

Perez, M.J., Beam, A.J., & Small, P.A. (2026). Race, memory, and colorblindness: Critical history and deconstructing United States democracy. *advances.in/psychology, 1*, e316437. https://doi.org/10.56296/aip00053

Pew Research Center. (2018, September 13). *Republicans, Democrats see opposing party as more ideological than their own*. https://www.pewresearch.org/politics/wp-content/uploads/sites/4/2018/09/9-13-2018-Party-Ideo-for-Release2.pdf

Pew Research Center. (2025, July 23). *Party affiliation fact sheet* (NPORS). https://www.pewresearch.org/politics/fact-sheet/party-affiliation-fact-sheet-npors/

Pickard, M. D., Roster, C. A., & Chen, Y. (2016). Revealing sensitive information in personal interviews: Is self-disclosure easier with humans or avatars and under what conditions? *Computers in Human Behavior*, *65*, 23–30. https://doi.org/10.1016/j.chb.2016.08.004

R Core Team. (2024). *R: A language and environment for statistical computing*. R Foundation for Statistical Computing. https://www.R-project.org/

Rambaud, S., Collange, J., Tavani, J. L., & Zenasni, F. (2021). Positive intergroup interdependence, prejudice, outgroup stereotype and helping behaviors: The role of group-based gratitude. *International Review of Social Psychology*, *34*(1), 10. https://doi.org/10.5334/irsp.433

Ren, S., Tomlinson, B., Black, R. W., & Torrance, A. W. (2024). Reconciling the contrasting narratives on the environmental impact of large language models. *Scientific Reports*, *14*(1). https://doi.org/10.1038/s41598-024-76682-6

Revelle, W. (2026). *Psych: Procedures for psychological, psychometric, and personality research* (R package version 2.6.1). Northwestern University. https://CRAN.R-project.org/package=psych

Roets, A., & Van Hiel, A. (2011). *Need for closure Scale--Short version*. PsycTESTS Dataset. https://doi.org/10.1037/t10237-000

Rosseel, Y. (2012). Lavaan: An R package for structural equation modeling. *Journal of Statistical Software*, *48*(2), 1–36. https://doi.org/10.18637/jss.v048.i02

Rosseel, Y., Jorgensen, T. D., & De Wilde, L. (2025). *Lavaan: Latent Variable Analysis* (R package version 0.6-21). https://doi.org/10.32614/CRAN.package.lavaan

Saguy, T., Tausch, N., Dovidio, J. F., & Pratto, F. (2009). The irony of harmony: Intergroup contact can produce false expectations for equality. *Psychological Science*, *20*(1), 114-121. https://doi.org/10.1111/j.1467-9280.2008.02261.x

Shah, J. Y., Kruglanski, A. W., & Thompson, E. P. (1998). Membership has its (epistemic) rewards: Need for closure effects on in-group bias. *Journal of Personality and Social Psychology, 75*(2), 383–393. https://doi.org/10.1037/0022-3514.75.2.383

Sheeran, P., & Webb, T. L. (2016). The intention-behavior gap. *Social and Personality Psychology Compass, 10*(9), 503–518. https://doi.org/10.1111/spc3.12265

Simchon, A., Edwards, M., & Lewandowsky, S. (2024). The persuasive effects of political microtargeting in the age of generative artificial intelligence. *PNAS Nexus*, *3*(2). https://doi.org/10.1093/pnasnexus/pgae035

Smith, E. R., Seger, C. R., & Mackie, D. M. (2007). Can emotions be truly group level? Evidence regarding four conceptual criteria. *Journal of Personality and Social Psychology*, *93*(3), 431–446. https://doi.org/10.1037/0022-3514.93.3.431

Spencer, S. J., Zanna, M. P., & Fong, G. T. (2005). Establishing a causal chain: Why experiments are often more effective than mediational analyses in examining psychological processes. *Journal of Personality and Social Psychology*, *89*(6), 845–851. https://doi.org/10.1037/0022-3514.89.6.845

Sundar, S. S. (2008). *The MAIN model: A heuristic approach to understanding technology effects on credibility* (pp. 73–100). Cambridge, MA: MacArthur Foundation Digital Media and Learning Initiative.

Sundar, S. S., & Kim, J. (2019, May). Machine heuristic: When we trust computers more than humans with our personal information. *Proceedings of the 2019 CHI conference on human factors in computing systems* (pp. 1–9).

Tajfel, H., & Turner, J. C. (1979). An integrative theory of intergroup conflict. In W. G. Austin & S. Worchel (Eds.), *The social psychology of intergroup relations* (pp. 33–48). Brooks/Cole.

Taniguchi, H. (2021). National identity, cosmopolitanism, and attitudes toward immigrants. *International Sociology*, *36*(6), 819–843. https://doi.org/10.1177/0268580921994517

Turner, J. C., Brown, R. J., & Tajfel, H. (1979). Social comparison and group interest in ingroup favouritism. *European Journal of Social Psychology*, *9*(2), 187–204. https://doi.org/10.1002/ejsp.2420090207

Udani, A., & Kimball, D. C. (2018). Immigrant resentment and voter fraud beliefs in the U.S. electorate. *American Politics Research*, *46*(3), 402–433. https://doi.org/10.1177/1532673X17722988

U.S. Census Bureau. (2024a). Age and sex. *American Community Survey, ACS 1-Year Estimates Subject Tables, Table S0101*. Retrieved December 27, 2025, from https://data.census.gov/table/ACSST1Y2024.S0101?q=Age+and+Sex+2024&y=2024.

U.S. Census Bureau. (2024b). *Educational attainment—People 18 years old and over by total money earnings in 2024, work experience in 2024, age, race, Hispanic origin, and sex* (Table PINC-04). Current Population Survey, 2025 Annual Social and Economic

Supplement. https://www.census.gov/data/tables/time-series/demo/income-poverty/cps-pinc/pinc-04.html

Valentino, N. A., Brader, T., & Jardina, A. E. (2013). Immigration opposition among U.S. Whites: General ethnocentrism or media priming of attitudes about latinos? *Political Psychology*, *34*(2), 149–166. https://doi.org/10.1111/j.1467-9221.2012.00928.x

Verkuyten, M. (2007). Religious group identification and inter-religious relations: A study among Turkish-Dutch Muslims. *Group Processes & Intergroup Relations*, *10*(3), 341–357. https://doi.org/10.1177/1368430207078695

Verkuyten, M. (2011). Assimilation ideology and outgroup attitudes among ethnic majority members. *Group Processes & Intergroup Relations*, *14*(6), 789–806. https://doi.org/10.1177/1368430211398506

Webster, D. M., & Kruglanski, A. W. (1994). Individual differences in Need for Cognitive Closure. *Journal of Personality and Social Psychology, 67*(6), 1049–1062. https://doi.org/10.1037/0022-3514.67.6.1049

Wenzel, M., Mummendey, A., & Waldzus, S. (2007). Superordinate identities and intergroup conflict: The ingroup projection model. *European Review of Social Psychology*, *18*(1), 331–372. https://doi.org/10.1080/10463280701728302

Wickham, H., Averick, M., Bryan, J., Chang, W., McGowan, L. D., François, R., Grolemund, G., Hayes, A., Henry, L., Hester, J., Kuhn, M., Pedersen, T. L., Miller, E., Bache, S. M., Müller, K., Ooms, J., Robinson, D., Seidel, D. P., Spinu, V., Yutani, H. (2019). Welcome to the tidyverse. *Journal of Open Source Software, 4*(43), 1686. https://doi.org/10.21105/joss.01686

Yetkili, O., Agdelen, N., Vural, S., & Kostyuk, E. (2025). The effects of identity (subordinate vs. superordinate) salience on intergroup attitudes, anxiety, and contact intentions in north Cyprus. *Peace and Conflict: Journal of Peace Psychology*, *31*(1), 26–35. https://doi.org/10.1037/pac0000747

Zarouali, B., Makhortykh, M., Bastian, M., & Araujo, T. (2021). Overcoming polarization with chatbot news? Investigating the impact of news content containing opposing views on agreement and credibility. *European Journal of Communication*, *36*(1), 53–68. https://doi.org/10.1177/0267323120940908

Zeileis, A., Köll, S., & Graham, N. (2020). Various versatile variances: An object-oriented implementation of clustered covariances in R. *Journal of Statistical Software*, *95*(1), 1–36. https://doi.org/10.18637/jss.v095.i01